\documentclass[conference]{IEEEtran}
\IEEEoverridecommandlockouts
\usepackage{cite}
\usepackage{amsmath,amssymb,amsfonts}
\usepackage{algorithmic}
\usepackage{lipsum}
\usepackage{graphicx}
\usepackage{comment}
\usepackage{enumitem}

\usepackage{booktabs}
\usepackage{balance}
\newtheorem{definition}{Definition}
\usepackage{algorithm}
\usepackage{algorithmic}
\usepackage{url}
\usepackage{bbm}
\usepackage{tabularx}
\ifCLASSOPTIONcompsoc
    \usepackage[caption=false, font=normalsize, labelfont=sf, textfont=sf]{subfig}
\else
\usepackage[caption=false, font=footnotesize]{subfig}
\fi

\usepackage{textcomp}
\usepackage{xcolor}
\def\BibTeX{{\rm B\kern-.05em{\sc i\kern-.025em b}\kern-.08em
    T\kern-.1667em\lower.7ex\hbox{E}\kern-.125emX}}
\begin{document}

\newcommand{\rev}[1]{\textcolor{black}{#1}}

\title{Cross-domain-aware Worker Selection with Training for Crowdsourced Annotation
}

\author{\IEEEauthorblockN{Yushi Sun\IEEEauthorrefmark{1}, Jiachuan Wang\IEEEauthorrefmark{1}, Peng Cheng\IEEEauthorrefmark{2}, Libin Zheng\IEEEauthorrefmark{3}, Lei Chen\IEEEauthorrefmark{4}\IEEEauthorrefmark{1}, Jian Yin\IEEEauthorrefmark{3}}
\IEEEauthorblockA{\IEEEauthorrefmark{1}\textit{Hong Kong University of Science and Technology}, Hong Kong, China}
\IEEEauthorblockA{\IEEEauthorrefmark{2}\textit{East China Normal University}, Shanghai, China}
\IEEEauthorblockA{\IEEEauthorrefmark{3}\textit{Sun Yat-sen University}, Guangzhou, China}
\IEEEauthorblockA{\IEEEauthorrefmark{4}\textit{Hong Kong University of Science and Technology (Guangzhou)}, Guangzhou, China}
\IEEEauthorblockA{ysunbp@cse.ust.hk, jwangey@connect.ust.hk, pcheng@sei.ecnu.edu.cn, \\ zhenglb6@mail.sysu.edu.cn, leichen@hkust-gz.edu.cn, issjyin@mail.sysu.edu.cn}}

\maketitle

\begin{abstract}
Annotation through crowdsourcing draws incremental attention, which relies on an effective selection scheme given a pool of workers. Existing methods propose to select workers based on their performance on tasks with ground truth, while two important points are missed. 1) The historical performances of workers in other tasks. In real-world scenarios, workers need to solve a new task whose correlation with previous tasks is not well-known before the training, which is called cross-domain. 2) The dynamic worker performance as workers will learn from the ground truth. In this paper, we consider both factors in designing an allocation scheme named cross-domain-aware worker selection with training approach. Our approach proposes two estimation modules to both statistically analyze the cross-domain correlation and simulate the learning gain of workers dynamically. A framework with a theoretical analysis of the worker elimination process is given. To validate the effectiveness of our methods, we collect two novel real-world datasets and generate synthetic datasets. The experiment results show that our method outperforms the baselines on both real-world and synthetic datasets. 
\end{abstract}

\begin{IEEEkeywords}
crowdsourcing, worker selection, cross-domain
\end{IEEEkeywords}

\section{Introduction} \label{sec:intro}

The quality of the labeled data is of great importance for the performance of machine learning, especially for supervised learning models~\cite{jain2020overview}. To get high-quality annotations for large-scale datasets, recruiting domain experts is too expensive and thus unacceptable. With a limited budget, annotation through selecting crowdsourcing workers is preferable and has drawn attention in recent years~\cite{shen2022optimization, cao2012whom}. Worker selection is one of the most important issues in the quality control consideration of crowdsourcing~\cite{zhao2017context}, which focuses on identifying workers with high performance from the worker pool. How to design an allocation scheme to effectively and efficiently select high-performance crowd workers remains a challenging problem.


\begin{figure}[t]
  \centering
  \includegraphics[width=\linewidth]{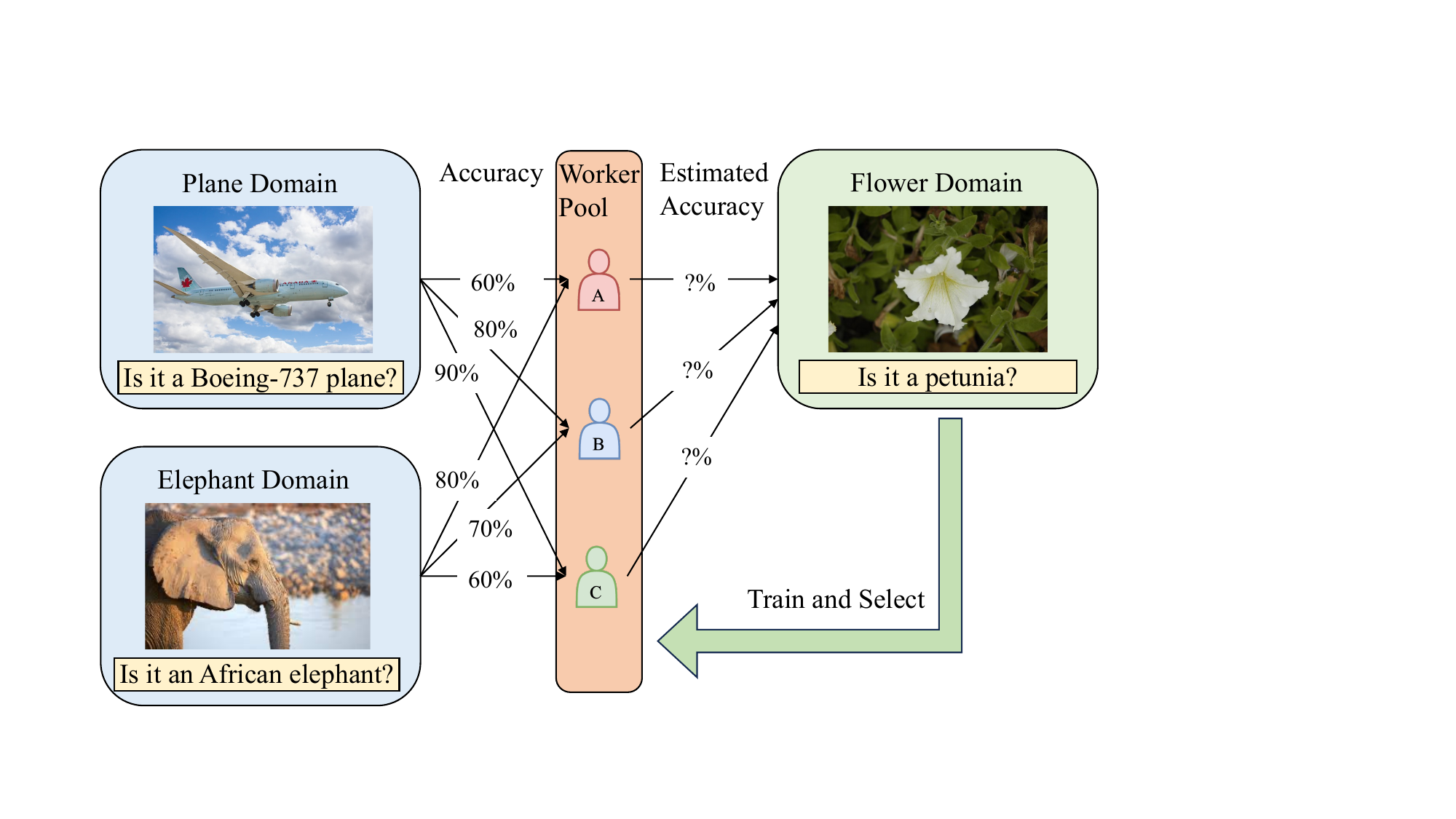}
  \vspace{-1em}
  \caption{Cross-domain worker selection. The left shows the two prior domains: plane and elephant. The right shows the target domain: flower. We record workers' historical accuracy on the two prior domains and estimate the accuracy on the target domain, to effectively train and select desired workers.}
  \label{fig:motivation}
  \vspace{-2em}
\end{figure}

In order to select workers through worker quality estimation, existing methods~\cite{li2015cheaper, zhao2017context, wu2021task, yadav2022multi} consider different factors in the crowdsourcing process: 1) workers' responses to the golden questions~\cite{li2015cheaper}; 2) additional social network interactions for worker trustworthiness estimation~\cite{zhao2017context, wu2021task}; 3) assumption of worker skills, which are hidden states between worker performance and tasks~\cite{yadav2022multi}.


{\color{black} In order to obtain large-scale manually labeled data for business, many companies such as JD.com, Inc.~\cite{JD-platform}, Alibaba~\cite{ali-platform}, and Baidu~\cite{baidu-platform} have their own commercial crowdsourcing platforms with worker pools.}
The answering history of workers stored in commercial crowdsourcing platforms can be helpful in selecting high-quality workers to complete tasks in a new domain~\cite{awwad2017efficient}, which is not well-explored in the existing worker selection methods~\cite{li2015cheaper, zhao2017context, wu2021task, yadav2022multi}. We refer to the tasks of new topics requiring workers to annotate as \textit{target domain tasks}, while the tasks of historical topics are  \textit{prior domain tasks}. In the beginning, the correlations between the target domain and these prior domains are not well-known, which is called \textit{cross-domain}. The performance of workers in the prior domain can help predict their performance in the target domain. As shown in Figure~\ref{fig:motivation}, given the classification performance on the elephants and planes of workers A, B, and C, we can obtain a rough idea of their domain knowledge of distinguishing living creatures and distinguishing machines, which is helpful for us to select the proper workers to work on tasks on other domains, such as flowers. Intuitively, workers with good performance in distinguishing elephants are likely to be sensitive to color and shape differences (since different kinds of elephants are similar in size but different in color and shape). In contrast, workers who perform well in distinguishing planes are likely to be good at identifying size differences (since different kinds of planes are similar in color and shape yet different in size). {\color{black}Given this prior domain knowledge, workers who are potentially good at distinguishing flowers (relying on color and shape differences) can be identified. We can train these identified workers by demonstrating the ground truth answers to them so that they can actively learn the characteristics of petunia and achieve better performance on the annotations. After that, we can select the best workers as the desired worker candidates for the target domain. In this way, the \textit{golden questions} from the target domain are fully utilized: not only used for estimating the cross-domain worker quality to select the best candidates but also used to boost the annotation performance of workers on the target domain through worker training.}



{However, transferring and incorporating workers' performance profiles across different domains is challenging.} Explicitly defining the mappings between the domains and the skill sets requires a comprehensive understanding of the domain tasks, which needs expert effort and thus fails to scale well in reality. Therefore, we propose automatically and inherently capturing the relationship between each domain and the required skills to ensure feasibility and scalability in real-world applications. Workers' performances are modeled based on reasonable assumptions for inner- and inter-domain. To be more specific, we apply normal distributions to model workers' performances on each domain following the modeling done by previous studies~\cite{shan2019crowdsourcing, zhao2015crowd} and adopt a multivariate normal distribution to model the correlation between workers' performances on different domains to achieve the goal of cross-domain worker selection.


As correlation is not well-known for the cross-domain problem, we apply a worker learning stage to train and select workers while simultaneously extracting their correlations. During the learning stage, limited \textit{golden questions} are given with accurate labels from experts. Previous approaches~\cite{liu2013scoring} treat the assignment of golden questions as a sampling process to get a static estimate of worker quality. However, workers' knowledge of the target domain can be dynamic~\cite{gadiraju2015training, haas2015argonaut}. For instance, in Figure~\ref{fig:motivation}, workers are asked whether the flower is a petunia. Initially, workers may have no idea what a petunia is. However, after we assign multiple golden (learning) questions regarding the petunia and reveal the answers to the workers, they can gradually learn about the characteristics (such as shape and color) of petunia and thus perform better when answering new questions on the same domain.

{Unfortunately, previous work has not studied the dynamic worker knowledge change in worker selection.} We fill the research gap by simulating workers as trainable to handle the dynamic worker selection instead of a static one. In this paper, we propose to model the learning gain of workers based on the Item Response Theory (IRT)~\cite{rasch1993probabilistic} from the Knowledge Tracing field to fully use the golden questions in selecting workers. Our allocation algorithm can select potential workers who can improve quickly during the learning stage, which are filtered out by static methods.

The contributions of this paper are as follows:
\begin{itemize}[leftmargin=*]

    \item We incorporate the cross-domain knowledge information and propose a novel Median Elimination-based worker selection with training algorithm to find high-quality workers.
    \item We comprehensively consider the learning gain of workers during the learning task worker training process over the new domain to get a better estimate of the dynamic change in worker quality.
    \item We collect two novel cross-domain worker selection datasets for the crowdsourcing research community to study the problem of cross-domain worker selection with training.
    \item We conduct extensive experiments on real-world and synthesized datasets to evaluate the performance of our proposed method comprehensively.
\end{itemize}

The following sections are arranged as follows. We first introduce the related work in Section~\ref{sec:related}. We then discuss the setup and problem formulation in Section~\ref{sec:setup}. 
The methodology is introduced in Section~\ref{sec:method}. We demonstrate the experiment results in Section~\ref{sec:exp}, and finally, we conclude the paper in Section~\ref{sec:conclude}.

\begin{figure}[t]
  \centering
  \includegraphics[width=\linewidth]{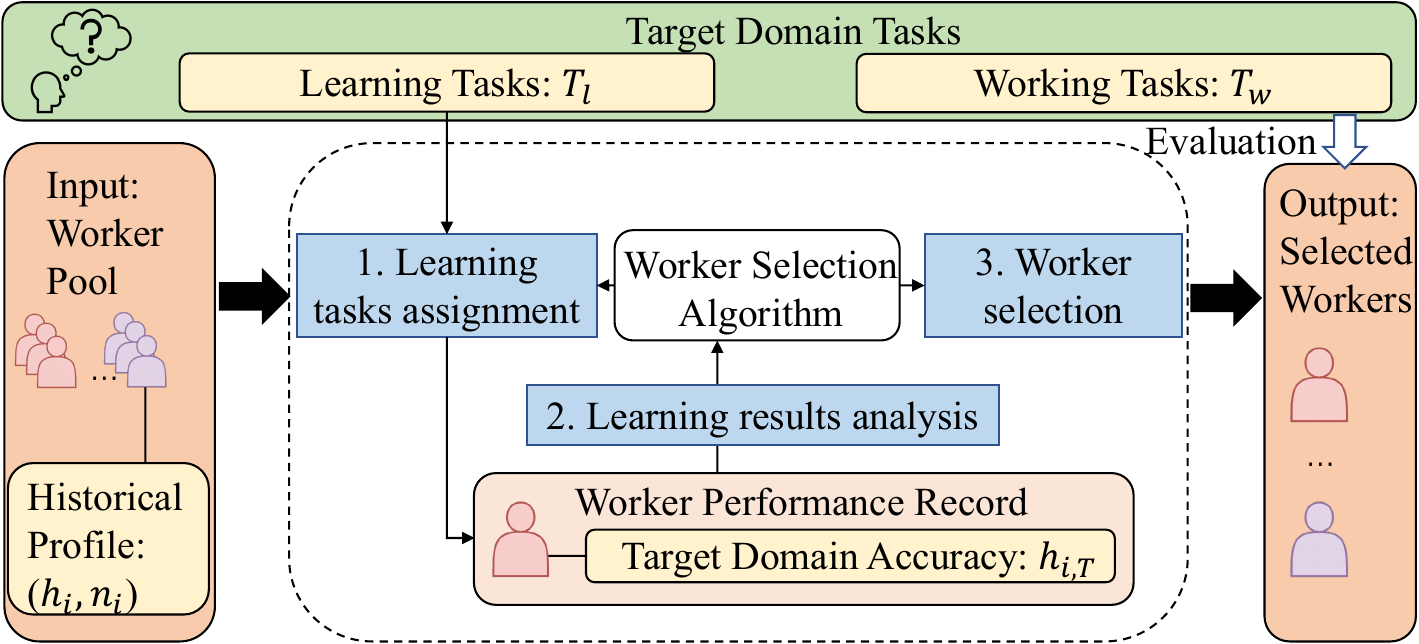}
  \caption{The definition of cross-domain-aware worker selection with training problem. The worker selection algorithm assigns learning tasks to workers, records and analyzes the learning task results, and performs worker selection. The performance of the selected workers is evaluated based on the target domain working tasks.}
  \label{fig:setup}
\vspace{-2em}
\end{figure}

\section{Related Work} \label{sec:related}

The general process of worker selection for quality control requires first estimating worker quality and then designing proper worker elimination algorithms to select the best workers. \rev{In this section, we first discuss the related works on worker selection from two aspects: worker quality estimation and worker elimination, which are the core components for worker selection with training task. Then we introduce the related works in Knowledge Tracing, which are related to the learning gain estimation process used for estimating worker quality and performing worker elimination. }

\subsection{Worker Quality Estimation}
Worker Quality Estimation assesses the abilities and reliability of individual workers participating in tasks on a crowdsourcing platform. Previous studies proposed various worker quality estimation methods based on the information available in different crowdsourcing worker selection scenarios. Liu et al.~\cite{ liu2013scoring} estimate worker quality based on workers' answers to golden questions. Li et al.~\cite{li2014wisdom} proposed a user discovery framework to select reliable workers based on general characteristics such as educational level, gender, and age. However, these characteristics may be too general for us to identify the best workers accurately. For instance, the approach would fail if the recruited workers were all college students with similar backgrounds. Zhao et al.~\cite{zhao2017context}, and Wu et al.~\cite{wu2021task} proposed integrating social network information into the worker quality estimation process. A major limitation of these approaches is that social network information is not necessarily available in many online crowdsourcing platforms, which limits the application domain of these approaches. Yadav et al.~\cite{yadav2022multi} proposed constructing a universal skill set with the mapping relationship between the skills and the crowdsourcing tasks. The skills of the workers are estimated based on the historical annotation performance. However, the universal skill set and the mapping relationship should be manually defined. Explicitly constructing the universal skill set and the mapping relationship between the skills and the tasks require domain knowledge, which brings overheads to real-world worker selection applications. \rev{Different from existing worker quality estimation approaches, we considered both cross-domain knowledge and the dynamic worker knowledge change to better capture the worker quality during the crowdsourcing process.}

\subsection{Worker Elimination}
This section introduces the worker elimination algorithms used by existing works. Worker elimination is a process used to select the most qualified workers for a task while filtering out underperforming or unreliable workers. Even-Dar et al.~\cite{even2006action} proposed a naive Uniform Sampling algorithm and a Median Elimination algorithm for the top-k selection multi-armed bandits problem, which can be adapted to perform worker elimination. Liu et al.~\cite{liu2013scoring} identify the best group of workers by assigning golden questions and selecting the workers that perform the best on those questions. Li et al.~\cite{li2014wisdom} selected the best workers based on the estimated performance generated from the general workers' profiles. Cao et al.~\cite{cao2015top} refined the theoretical bounds of~\cite{even2006action}. They introduced budget constraints and proposed a greedy-based heuristic algorithm to sort the workers based on the error rate and the requirements. Zhao et al.~\cite{zhao2017context} proposed forward and backward selection algorithms based on social network connections to gradually identify the best workers. Wu et al.~\cite{wu2021task} considered the interest similarity between the workers and the tasks based on social network information to identify the best workers for the tasks. Yadav et al.~\cite{yadav2022multi} proposed a team formation algorithm to gather the workers with the desired expertise for the target tasks. \rev{Building on the Medium Elimination algorithm introduced by~\cite{cao2015top}, we additionally considered the worker learning gain during the worker elimination process, so as to achieve better worker elimination results.}

\subsection{Knowledge Tracing}
Knowledge tracing is a technique used in education to understand how well a student is learning a particular subject. It involves tracking and predicting a student's knowledge and understanding over time. By analyzing the student's responses to questions or tasks, knowledge tracing models can estimate the student's current level of knowledge, identify areas of strength and weakness, and provide personalized feedback and guidance to enhance learning~\cite{corbett1994knowledge}. As discussed in~\cite{abdelrahman2022knowledge, minn2018deep}, the knowledge tracing methods can be divided into three categories: Bayesian Knowledge Tracing, Factor Analysis Models, and Deep Knowledge Tracing based on the different types of inputs and application scenarios.

\noindent \textit{Bayesian Knowledge Tracing (BKT): }The BKT model is first introduced by~\cite{corbett1994knowledge}, where the skills behind the questions are considered. Four different types of probabilities associated with changes in skill mastery are modeled, and the Bayesian estimation of the final state skill mastery probability is used. Several subsequent works~\cite{khajah2014integrating, lee2012impact, yudelson2013individualized, d2008more, pardos2011kt} propose to extend the original BKT model with student-specific modeling and inter-skill relationship.

\noindent \textit{Factor Analysis Models: }The simplest and the most widely used Item Response Theory (IRT) model is Rasch's model~\cite{rasch1993probabilistic}, which defines a one-parameter logistic (1PL) IRT model. The probability that a worker answers a question correctly is modeled as a logistic function based on the worker's learning parameters and the difficulty of the question. Wilson et al.~\cite{wilson2016back} extended the original 1PL IRT model to Hierarchical IRT and Temporal IRT by additionally considering the relatedness of parameters across different questions and times, respectively. Performance Factor Analysis (PFA)~\cite{pavlik2009performance} is proposed to extend the IRT model by replacing the learning parameter with multiple learning skill parameters to model the relationship of multiple skills.

\noindent \textit{Deep Knowledge Tracing (DKT): }DKT is first proposed in~\cite{piech2015deep}, which models the knowledge states (skills) of people with Long Short Term Memory (LSTM)~\cite{hochreiter1997long}. The LSTM network contains many neurons to represent the hidden states of workers' answer history. The current knowledge states of workers can be learned from the training data. Several works~\cite{zhang2017dynamic, miller2016key, abdelrahman2019knowledge} extend the idea of the original DKT model to achieve improved performance.

In our paper, we aim to model the worker learning gain without explicitly defining and modeling the relationship between the skills and the questions, so we adopt Rasch's IRT model~\cite{rasch1993probabilistic} to model the learning process of workers while answering learning questions. \rev{Note that the focus of this paper is to introduce the knowledge tracing techniques into the crowdsourcing worker selection process to achieve better worker quality estimation and elimination results, instead of developing novel knowledge tracing approaches.}

\section{Problem Formulation} \label{sec:setup}

\begin{table}[t!]
  \caption{Notations.}
\vspace{-1em}
  \begin{center}
  \begin{tabularx}{\columnwidth}{p{1.3 cm} m{6.5 cm}}
    \hline
    \textbf{Notations} & \textbf{Descriptions}\\
    \hline
    $T_l$, $T_w$ & learning tasks and working tasks sets\\
    $W$ & the worker pool\\
    $w_i$ & the $i$-th worker in the worker pool $W$\\
    \rev{$h_i$} & \rev{the historical profile of the worker $w_i$}\\
    \rev{$n_i$} & \rev{number of annotation tasks completed by worker $w_i$ on different domains}\\
    $B$ & the total budget\\
    $k$ & number of workers we want to select\\
    $\alpha_i$ & the learning parameter of worker $w_i$\\
    $\beta_d$ & the domain difficulty parameter for domain $d$\\
    \rev{$\theta_i$} & \rev{the proficiency parameter of worker $w_i$}\\
    \rev{$K_j$} & \rev{the cumulative number of learning tasks assigned to each remaining worker till round $j$}\\
    $n$ & the number of elimination rounds\\
    $a_t$ & the initialized annotation accuracy of the target domain\\
    $Q$ & the number of learning tasks per batch\\
    \hline
\end{tabularx}
\end{center}
\vspace{-2em}
\label{tab:notation}

\end{table}

We present the notations used in this paper in Table~\ref{tab:notation}. The general process of cross-domain-aware worker selection with training can be divided into three steps, as shown in Figure~\ref{fig:setup}. We first discuss the tasks and workers considered in our paper, introduce the three steps generally, and formally define the cross-domain-aware worker selection with training problem.

\begin{definition} {\rm(Tasks)}. The crowdsourcing tasks on the target domain can be categorized into learning and working tasks. The learning tasks (golden questions) refer to the tasks that have gold labels. The working tasks are the tasks without gold labels. We denote the set of tasks on the target domain as $T$, the set of learning tasks as $T_l$, and the set of working tasks as $T_w$.
\end{definition}
 For simplicity, we consider Multiple Choice Question tasks in our paper. As suggested by~\cite{awwad2017efficient}, this selection of task type does not influence the generalizability of our approach since our approach is based on the answering accuracy, which can also be computed if other kinds of tasks are used.

\begin{definition} {\rm(Workers)}. We denote the worker pool as $W$. Each worker $w_i$ in $W$ is associated with a historical profile $(h_i, n_i) = (\{h_{i,1}, h_{i,2}, ..., h_{i,D}\}, \{n_{i,1}, n_{i,2}, ..., n_{i,D}\})$ where $D$ is the number of prior domains, $h_{i,j}$ is the annotation accuracy of worker $w_i$ on the $j$-th prior domain, and $n_{i,j}$ is the number of annotation tasks completed by worker $w_i$ on the $j$-th prior domain. The annotation accuracy of $w_i$ on the target domain working tasks is denoted as $h_{i,T}$.
\end{definition}

\begin{definition} {\rm(Learning tasks assignment)}. Learning tasks assignment is the process of assigning learning tasks to the worker and recording the accuracy of each worker. The answers to the learning tasks are revealed to each worker after he/she submits the answers so that he/she can learn the target domain knowledge from the revealed ground truths.
\end{definition}

In our paper, we train workers in rounds. Each remaining worker $w_i$ is assigned learning tasks and the annotation accuracy $a_{i,c}$ is recorded in round $c$.

\begin{definition}
    {\rm(Learning results analysis)}. Learning results analysis is the process of estimating workers' performances based on their results on the learning tasks for the assignment of remaining learning tasks and the selection of workers.
\end{definition}

\begin{definition}
    {\rm(Worker selection)}. Considering the cross-domain performances and learning result feedback of workers, the worker selection step is to select the well-trained workers with the best performance in the target domain.
\end{definition}



\begin{definition}
    {\rm (Cross-domain-aware worker selection with training)}. Given target domain tasks $T = \{T_l, T_w\}$, the total budget $B$, and worker pool $W$ with each worker $w_i$'s historical profile $h_i$. Cross-domain-aware worker selection with training problem is to 1) properly assign no more than $B$ tasks to $|W|$ workers for training based on workers' historical records and learning feedback and 2) select top k workers with the highest possible annotation accuracy on working tasks $T_w$.
    
\end{definition}

\begin{figure*}[t]
  \centering
  \includegraphics[width=0.9\linewidth]{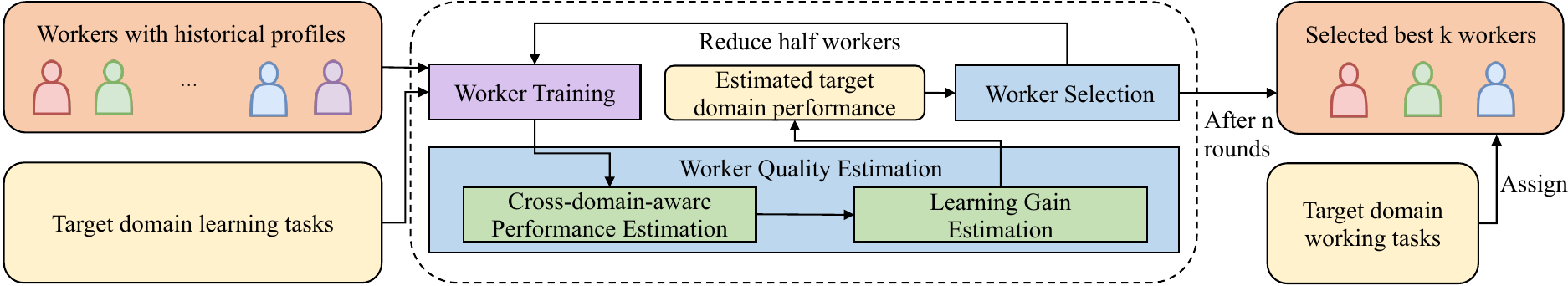}
  \vspace{-1em}
  \caption{The general pipeline of our cross-domain-aware worker selection with training algorithm. }
  \label{fig:pipeline}
  \vspace{-2em}
\end{figure*}

\section{Methodology} \label{sec:method}

In this section, we first introduce our general framework for cross-domain-aware worker selection with training problem (Subsection~\ref{subsec:framework}). Then we demonstrate the process of worker training (Subsection~\ref{subsec:worker_training}) and discuss the details of the two core phases: Worker Quality Estimation (Subsection~\ref{sec:worker_quality_estimation}) and Worker Selection (Subsection~\ref{sec:worker_selection}) with theoretical analysis. 
A summary of the whole pipeline is presented in Subsection~\ref{sec:summary}.

{\color{black}\subsection{Framework}\label{subsec:framework}

We display our framework in Figure~\ref{fig:pipeline}. Workers are iteratively trained and selected through a 3-phase pipeline:
\begin{itemize}[leftmargin=*]
    \item Worker Training. In the target domain, workers answer questions and check answers to renew their knowledge.
    \item Worker Quality Estimation. The ability estimation of each worker will be updated according to his/her answers during worker training in addition to the historical records on target and other domains.
    \item Worker Selection. Based on the estimated worker quality, we select the best half of the workers to enter the next round.
\end{itemize}
Finally, after $n$ rounds, we obtain the selected best $k$ workers and assign the target domain working tasks for them to annotate. 
In our paper, we fix the budgets in each round and focus on the accuracy of dynamic worker estimation. \rev{Mathematically, $t=\lfloor \frac{B}{n} \rfloor$ is the fixed number of learning tasks per round, and $|W_c|$ is the number of remaining workers for the current round $c$. Then, $\lfloor \frac{t}{|W_c|} \rfloor$ is the number of tasks per worker for round $c$.} 

{\color{black}\subsection{Worker Training}\label{subsec:worker_training}
We can summarize the worker training as a simple ``Answer and Learn'' process for workers. To be more specific, after a worker completes one batch of learning tasks, their ground truth answers will be revealed to the worker. For example, as shown in Figure~\ref{fig:problem}, the left shows a learning task completed by a worker, and the right shows the ground truth answer to that task. A worker can learn from the ground truth answers and renew his/her target domain knowledge. 

Formally, after each round of task assignment, we denote the answers of worker $w_i$ in the current round $c$ as $a_{i,c}$. 
}


\subsection{Worker Quality Estimation} \label{sec:worker_quality_estimation}

{\color{black}To achieve high-quality worker selection, two factors are important: \textit{cross-domain correlation} which can help us filter workers according to their performance on other domains; and \textit{worker learning gain} where workers who improve more from training should be preserved and assigned with more training tasks.}
Thus, we divide the worker quality estimation phase into Cross-domain-aware Performance Estimation (CPE) and Learning Gain Estimation (LGE). CPE focuses on modeling the cross-domain correlation of workers, while LGE focuses on modeling the learning gain in the worker training process.

\subsubsection{Cross-domain-aware Performance Estimation} 


As stated in Section~\ref{sec:intro}, cross-domain information is important for worker selection. However, no correlation information between the prior domains and the target domain is available before worker training. Instead, worker feedback on learning tasks is accumulatively arriving, which requires a mining algorithm to dynamically capture the cross-domain correlation from scratch. In this subsection, we introduce our CPE estimation scheme, which 1) derives statistically analyzed expectation of accuracy and 2) supports online updates with a Maximum Likelihood Estimation as the base. We present the whole CPE estimation in Algorithm~\ref{algo:MLE}.

In order to model the correlation between workers' prior domain knowledge and the target domain knowledge, we adopt the multivariate normal distribution~\cite{tong2012multivariate}.  Precisely, to model the relationship between the $D$ prior domains and the target domain effectively, we adopt a $(D+1)$-dimensional multivariate normal distribution $\mathcal{N}(\mu, \Sigma)$, where $\mu \in \mathbb{R}^{(D+1)}$ and $\Sigma \in \mathbb{R}^{(D+1) \times (D+1)}$:

\vspace{-1em}
    \begin{gather} \label{eq:initha}
\mu = [\mu_1, \mu_2,..., \mu_{D}, \mu_{T}]^\mathsf{T} ,\\
\Sigma = \begin{bmatrix}
    \sigma_1^2&\rho_{1,2}\sigma_1\sigma_2& ... &\rho_{1,D}\sigma_1\sigma_D&\rho_{1,T}\sigma_1\sigma_T\\
    \rho_{2,1}\sigma_2\sigma_1&\sigma_2^2& ... &\rho_{2,D}\sigma_2\sigma_D&\rho_{2,T}\sigma_2\sigma_T\\
    ... & ... & ... & ... & ...\\
    \rho_{D,1}\sigma_D\sigma_1&\rho_{D,2}\sigma_D\sigma_2& ... &\sigma_D^2&\rho_{D,T}\sigma_D\sigma_T\\
    \rho_{T,1}\sigma_T\sigma_1&\rho_{T,2}\sigma_T\sigma_2& ... &\rho_{T,D}\sigma_T\sigma_D&\sigma_T^2\\
        \end{bmatrix}\label{eq:init} ,
\end{gather}
\noindent the $\mu_i, \sigma_i, \rho_{i,j}$ where $i \neq j$ and $i,j \in \{1,2,..., D, T\}$ are the mean and standard deviation of worker accuracy on each domain and the correlation parameters between any pair of domains respectively.

The annotation accuracy for each worker $w_i$ on each domain is modeled as a $(D+1)$-dimensional random vector $v_i = [h_{i,1}, h_{i,2}, ..., h_{i,D}, h_{i,T}]^\mathsf{T} \in \mathbb{R}^{(D+1)}$, where $v_i \sim \mathcal{N}(\mu, \Sigma)$. 

As shown in Figure~\ref{fig:pipeline}, we perform CPE (Algorithm~\ref{algo:MLE}) in each elimination round. Specifically, each worker is assigned $(\lfloor \frac{t}{|W_c|} \rfloor)$ learning tasks and we record the answers for each worker $w_i$: $a_{i,c}=[a_{i,c,1},a_{i,c,2},...,a_{i,c,\lfloor \frac{t}{|W_c|} \rfloor}]$ and store to $A_c$. For each worker, we compute the number of correct and wrong answers as follows:
    \begin{gather} 
    C_{i,c} = \sum_{j=1}^{\lfloor t/|W_c| \rfloor} \mathbbm{1}(a_{i,c,j} = g_{j,c}) \label{eq:teq}, \\
    X_{i,c} = \lfloor (t/|W_c|) \rfloor - C_{i,c} .\label{eq:feq}
\end{gather}
\indent Given each worker's correct and wrong answers in each round, we adopt Maximum Likelihood Estimation to estimate $\mu$ and $\Sigma$. The log-likelihood function $L$ is formulated as follows:
    \begin{gather}
\begin{aligned} 
\begin{split}
    \log{L} =& \sum_{i=1}^{|W_c|} \log{P(h_{i,T}|h_i)} \\
      =& \sum_{i=1}^{|W_c|} \log{\int_{0}^{1} h_{i,T}^{C_{i,c}} (1-h_{i,T})^{X_{i,c}} \frac{e^{-\Psi}}{\sqrt{2\pi |\Bar{\Sigma}|}}} \mathrm{d} h_{i,T}\\
     = & \sum_{i=1}^{|W_c|} \big[\log{\int_{0}^{1} h_{i,T}^{C_{i,c}} (1-h_{i,T})^{X_{i,c}} e^{-\Psi}}\mathrm{d} h_{i,T}\\& +\log{\frac{1}{\sqrt{2\pi}}}-\frac{1}{2}\log{|\Bar{\Sigma}|}\big],\label{eq:MLE}
\end{split}
\end{aligned}
\end{gather}
where $\Bar{\mu}$ and $\Bar{\Sigma}$ are the conditional distribution of the multivariate normal distribution $\mathcal{N}(\mu, \Sigma)$:
    \begin{gather}
    \Bar{\mu} = \mu_T+\Sigma_{1\times D}\Sigma_{D\times D}^{-1}(h_i-\mu_{1\sim D})\nonumber ,\\
\Bar{\Sigma} = \Sigma_{1\times 1}-\Sigma_{1\times D}\Sigma_{D\times D}^{-1}\Sigma_{D\times 1},\nonumber
\end{gather}
{\color{black}and $\Psi=\frac{(h_{i,T}-\Bar{\mu})^\mathsf{T}(h_{i,T}-\Bar{\mu})}{2\Bar{\Sigma}}$.

In real-world applications, these parameters are updated under a large amount of streaming data, where directly calculating the optimal parameters is unacceptable. To enable an incremental parameter estimation,} we update $\mu$ and $\Sigma$ by maximizing Equation~\eqref{eq:MLE} with gradient descent in each round:
    \begin{gather}
    \rev{\mu' = \mu- r_1 \nabla_\mu \log L},\\
    \rev{\Sigma' = \Sigma- r_2 \nabla_\Sigma \log L},
\end{gather}
\rev{where $r_1$ and $r_2$ are the learning rates of gradient descent for $\mu$ and $\Sigma$; $\mu'$ and $\Sigma'$ are the updated $\mu$ and $\Sigma$ at the current gradient descent step.} After obtaining the Maximum Likelihood Estimation of the mean $\hat{\mu}$ and standard deviation $\hat{\Sigma}$, we obtain the predicted annotation accuracy for each worker $w_i$ with the updated multivariate normal distribution $\mathcal{\hat{N}}(\hat{\mu}, \hat{\Sigma})$:
\begin{equation}
\label{eq:p4}
    \begin{aligned} 
    p_{c,i} &= E[h_{i,T}|h_i]\\
            &= \int_{0}^{1} h_{i,T} P(h_{i,T}|h_i) \mathrm{d} h_{i,T}\\
            &= \int_{0}^{1} h_{i,T} \frac{P(h_i, h_{i,T})}{P(h_i)} \mathrm{d} h_{i,T}, \\
    \end{aligned}
\end{equation}
\noindent where $[h_i, h_{i,T}]^\mathsf{T} \sim \hat{\mathcal{N}}(\hat{\mu}, \hat{\Sigma})$ and $[h_i]^\mathsf{T} \sim \hat{\mathcal{N}}(\hat{\mu}_{1 \sim D}, \hat{\Sigma}_{D \times D})$. $p_{c,i}$ is used as the estimated worker accuracy in the target domain instead of a coarse observation on $C_{i,c}$ and $X_{i,c}$.

\begin{figure}[t]
  \centering
  \includegraphics[width=0.75\linewidth]{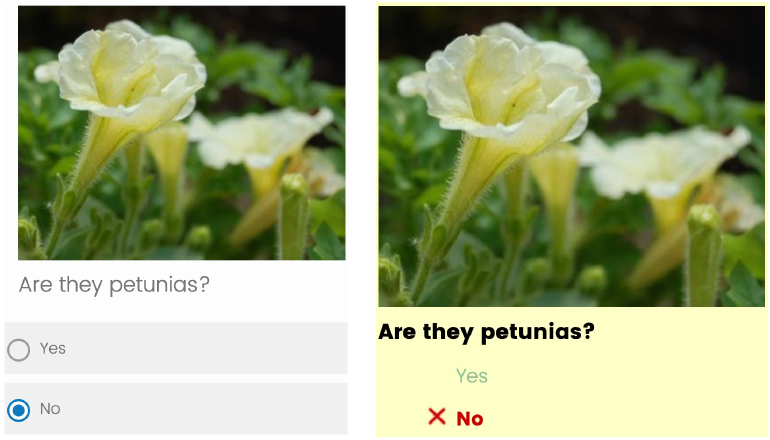}
  \caption{An illustration of the learning task (left) and its corresponding ground truth answer (right). The learning tasks will be displayed to the workers. After they complete their current round answers, the ground truth will be revealed for them to learn.}
  \label{fig:problem}
\end{figure}

\begin{algorithm}[t!]
\caption{Cross-domain-aware Performance Estimation (CPE)}
\label{algo:MLE}
\begin{algorithmic}[1]
\REQUIRE ~~\\
The answers of workers in current round $A_c$\\
The learning tasks ground truth in current round $G_c$\\
The historical accuracy of workers in current round $H_c$\\
The number of workers remaining in current round $|W_c|$
\ENSURE ~~\\
The predicted accuracy of remained workers $p_c$\\
\STATE Initialize $p_c$ to be an empty array
\STATE Initialize the multivariate normal distribution $\mathcal{N}(\mu, \Sigma)$ 
\STATE Compute the number of correct and wrong tasks of each worker $C_{i,c}, X_{i,c}$ according to Equations~\eqref{eq:teq} and \eqref{eq:feq} 
\STATE Compute updated distribution $ \hat{\mathcal{N}}(\hat{\mu}, \hat{\Sigma})$, where $$\hat{\mu}, \hat{\Sigma} = \arg\max_{\mu, \Sigma} \log L(\mathcal{N}, \left\{C_{i,c}, X_{i,c}\right\}_{i=1}^{|W_c|})$$
\FOR{each $h_i$ in $H_c$}
\STATE Compute $p_{c,i}$ via Equation~\eqref{eq:p4} and append to $p_c$
\ENDFOR
\RETURN $p_c$
\end{algorithmic}
\end{algorithm}

\subsubsection{Learning Gain Estimation} 

Through the interactive mode displayed in Figure~\ref{fig:problem}, workers not only provide feedback but also \textit{learn} from the results. This training process plays an important role in crowdsourcing~\cite{gadiraju2015training, haas2015argonaut}, but is hardly studied together with worker selection. This motivates us to enhance worker estimation with training, which aims at capturing the changes in estimated target domain performance for workers. This further enables us to quantize each worker's learning gain after assigning a certain amount of learning tasks to get a more accurate dynamic estimation of workers' performance on $T_w$.

In order to capture the learning gains, we adopt the item response theory model used for capturing the student learning process from the previous work~\cite{wilson2016back,minn2018deep}. In the original IRT model~\cite{wilson2016back,minn2018deep}, considering a worker $w_i$, the probability that $w_i$ answers question $q$ from domain $d$ correctly is:
\begin{equation} 
    p_d(\theta_i)=(1+e^{-(\theta_i-\beta_d)})^{-1} ,\label{eq:IRT}
\end{equation}
\noindent where $\theta_i$ is the proficiency parameter of the worker and $\beta_d$ is the difficulty parameter of the question $q$ from domain $d$. \rev{In our setting, $\theta_i=\alpha_i \ln(K_j+1)$, which is proportional to the logarithm of the cumulative number of learning tasks ($K_j=\frac{(2^j-1)*t}{|W|}$ for the target domain) assigned to worker $w_i$, while $\alpha_i$ is computed through least square regression and will be discussed in Equation~\eqref{eq:lsr_obj}.}
Different difficulty parameters are assigned to tasks in different domains, denoted as $\beta_{1 \sim D}$ for tasks in prior domains and $\beta_T$ for tasks in the target domain. The modified item response theory model for a single worker $w_i$ at the learning stage $j$ on the domain $d$ is:
\begin{equation} 
\begin{aligned} \label{eq:IRTmodi}
    \hat{p}_{j,i,d} &= g(\alpha_i, \beta_d, K_j)\\
                  &= \frac{1}{1+e^{-(\alpha_i \ln(K_j+1) -\beta_d)}}.
\end{aligned}
\end{equation}

The last step before one can further estimate the dynamic performance after training instead of a static performance is to get \rev{the update formula for }the intrinsic learning parameter $\alpha_i$ of each worker \rev{in each round}. \rev{We update the learning parameter $\alpha_i$ by minimizing the following least square regression objective:}
\begin{equation}
    \begin{aligned}
    \alpha_i = 
    \arg\min_{\alpha_i} \bigg[\sum_{d=1}^{D} (\hat{p}_{1,i,d}-h_{i,d})^2+\sum_{j=1}^{c} (\hat{p}_{j-1,i,t}-p_{j,i})^2\bigg],
    \label{eq:lsr_obj}
    \end{aligned}
\end{equation}
\noindent which comprises two parts: the first part minimizes gaps between learning gain estimations and accuracy on $D$ prior domains; the second part minimizes each pair of worker accuracy estimated by Equation~\eqref{eq:IRTmodi} in round $j-1$ and CPE in round $j$. The round index is different because the CPE estimation is based on the workers' performances in round $j$, where workers are only shown with $j-1$ rounds of answers (trained with $j-1$ rounds). \rev{The regression is conducted in each round to update each $\alpha_i$. }

According to the results of each round (e.g., the $j^{th}$ round), we assign tasks to workers and expect the best performance after the training of the next round (e.g., the $j+1^{th}$ round), which can be obtained through Equation~\eqref{eq:IRTmodi} (e.g., compute $\hat{p}_{j+1,i,d}$), which is intractable for static methods. We argue that such an estimation is closer to the actual annotation performance on the working tasks after $n$ rounds of training and thus can help us get a more accurate estimate of the actual value of $h_{i,T}$ for each worker $w_i$. {\color{black}We display the LGE in Algorithm~\ref{algo:LPE}. }


\begin{algorithm}[t!]
\caption{Learning Gain Estimation (LGE)}
\label{algo:LPE}
\begin{algorithmic}[1]
\REQUIRE ~~\\
The workers remained in the current round $W_{c}$\\
The historical accuracy of workers in the current round $H_c$\\
The historical task numbers of workers in the current round $N_c$\\
The predicted accuracy arrays $p_1, p_2, ..., p_c$ at the current stage
\ENSURE ~~\\
The updated predicted accuracy with learning gains $\hat{p}_c$ 
\STATE Initialize $\hat{p}_c$ as an empty array
\STATE Initialize the difficulty parameters in the target domain ($\beta_T$) and prior domains ($\beta_1, \beta_2,\cdots, \beta_D$)
\FOR{each $w_i \in W_{c}$}
\STATE Initialize the learning parameter $\alpha_i$
\FOR{domain $d = 1, 2, ..., D$}
\STATE Compute the historical accuracy  $h_{i,d}$
\STATE Compute the historical task numbers $n_{i,d}$
\STATE Compute the IRT score $\hat{p}_{1,i,d}=g(\alpha_i, \beta_d, n_{i,d})$
\ENDFOR
\FOR{stage $j=1,2,...,c$}
\STATE Compute the predicted accuracy for $w_i$: $p_{j,i}$
\STATE Compute $\hat{p}_{j-1,i,T} = g(\alpha_i, \beta_T, K_{j-1})$
\ENDFOR
\STATE \rev{Update $\alpha_i$ according to Equation~\eqref{eq:lsr_obj}}
\STATE Compute $\hat{p}_{c,i,T} = g(\alpha_i, \beta_T, K_c)$
\STATE Append $\hat{p}_{c,i,T}$ to $\hat{p}_c$
\ENDFOR
\RETURN $\hat{p}_c$
\end{algorithmic}
\end{algorithm}

\subsection{Worker Selection} \label{sec:worker_selection}
{\color{black}Based on the above estimations, we propose our algorithm for worker selection in this subsection, where a theoretical guarantee is given. }Compared with the intuitive but effective design of the Median Elimination algorithm discussed in~\cite{even2006action}, we have a fixed amount of budget to allocate tasks, where the original algorithm and theory cannot be directly applied. Here, we propose an adaptation version, displayed in Algorithm~\ref{algo:ME}. To be more specific, ME is called in rounds, where in each round, the worst half of workers are eliminated. The algorithm terminated with k workers left as its output. With a limited budget, we reversely derive the number of rounds needed for elimination and allocate the budget. Specifically, given worker pool $W$, number of $k$, total budget $B$, we can get the total round $n$ and the budget allocated in each round $t$ as:
    \begin{gather}
    n = \lceil \log (|W|/k) \rceil \label{eq:n} ,\\
t = \lfloor B/n \rfloor \label{eq:t} .
\end{gather}
\indent Unlike the original $(\epsilon, \delta)$ bound formulation in~\cite{even2006action}, we constrain the total budget used for the task and prove a theoretical bound over error $\epsilon_c$ in each round. Specifically, given a fixed total budget of $B$, the algorithm aims at finding the top k workers. It satisfies that the best worker outputted by the algorithm in round $c+1$ is an $\epsilon_c$-optimal worker with respect to the best worker outputted in round $c$, with a probability of least $1-\delta_c$. The error at each round $c$ is bounded by $O(\sqrt{(\frac{nk}{B})\ln{(\frac{1}{\delta_c})}})$. 

Adapting from the proof of Lemma 11 in~\cite{even2006action}, we have the following theoretical results:

\noindent \textbf{Theorem 1} \textit{By applying our adapted ME algorithm, we have:}
\begin{equation} 
    P[\max_{w_j \in W_c} h_{j, T} \le \max_{w_i \in W_{c+1}} h_{i, T} + \epsilon_c ] \ge 1-\delta_c,
\end{equation}
where each worker is assigned $(\frac{2}{\epsilon_c^2})\ln{( \frac{3}{\delta_c})}$ tasks in round $c$.

\noindent \textbf{Proof} We present the proof in our technical report~\cite{technical-report}.

According to the above theorem,  we can derive the following bound for our allocation scheme:

\noindent \textbf{Theorem 2} \textit{In each round $c$ of Algorithm~\ref{algo:general}, the error $\epsilon_c$ is bounded by $O(\sqrt{(\frac{nk}{B})\ln{(\frac{1}{\delta_c})}})$.}

\noindent \textbf{Proof} We present the proof in our technical report~\cite{technical-report}.

\subsection{Summary} \label{sec:summary}
The algorithm regarding the whole pipeline is summarized in Algorithm~\ref{algo:general}. Workers with historical profiles (prior domain performance) are first assigned target domain learning tasks for training purposes (Line 9 of Algorithm~\ref{algo:general}). The annotation accuracy is recorded, then we enter the worker quality estimation phase (Lines 13-14 of Algorithm~\ref{algo:general}): we perform Cross-domain-aware Performance Estimation to generate an estimation of the worker accuracy, and we further use Learning Gain Estimation to estimate the performance gains. Finally, we perform worker selection (Line 15 of Algorithm~\ref{algo:general}) by applying Median Elimination in each round to select the best half of workers. After $n$ rounds of looping, we obtain the selected best $k$ workers on the target domain tasks. \rev{As for the time complexity, as shown in Algorithm~\ref{algo:general}, we have $n$ iterations. Let $G$ be the number of gradient descent epochs performed to maximize Equation~\ref{eq:MLE}. In each iteration, we perform CPE, LGE, and ME, which take $O(G|W_c|)$, $O(|W_c|\log(|W_c|/k))$, and $O(|W_c|\log(|W_c|))$, respectively. Therefore, the overall time complexity for the worker quality estimation and selection process is $O(n|W|(G+\log(|W|)))$. We do not consider the time for workers to complete the learning tasks when analyzing the time complexity and we will discuss this in Section~\ref{sec:runtime}.}

{\color{black}Note that our solution is not restricted to the case where workers have been working on all the $D$ domains. For each domain $d$, if worker $w_i$ does not have historical record $h_{i, d}$, we can remove the corresponding $d^{th}$ row and line in $\mu$ and $\sum$ of Equation~\eqref{eq:MLE} for worker $w_i$ and remove the addition term $(\hat{p}_{1,i,d}-h_{i,d})^2$ in Equation~\eqref{eq:lsr_obj}, so that our approach can still work if any workers have not been working on all the prior domains. In this way, one can easily adapt our approach to suit the general cases.}
\begin{algorithm}[t!]
\caption{Median Elimination (ME)}
\label{algo:ME}
\begin{algorithmic}[1]
\REQUIRE ~~\\
The predicted accuracy $\hat{p}_c$\\
The workers remained in the current round $W_c$
\ENSURE ~~\\
The selected workers $W_{c+1}$\\
\STATE $w_1, w_2,..., w_{|W_c|} = $ the workers sorted in non-increasing order of their predicted accuracy $\hat{p}_c$
\STATE $W_{c+1} = \{w_1, w_2, ..., w_{\lceil \frac{|W_c|}{2}\rceil}  \}$
\RETURN $W_{c+1}$
\end{algorithmic}
\end{algorithm}

\begin{algorithm}[t]
\caption{General Algorithm}
\label{algo:general}
\begin{algorithmic}[1]
\REQUIRE ~~\\
A set of workers $w_i \in W$, a set of learning tasks $t_j \in T_l$\\
Workers' historical accuracy $h_i=\{h_{i,1}, h_{i,2}, ..., h_{i,D}\}$\\
Workers' historical task number $n_i=\{n_{i,1}, n_{i,2}, ..., n_{i,D}\}$\\
Total budget $B$\\
Probability $\delta$
\ENSURE ~~\\
The set of selected top $k$ workers $W_T$\\
\STATE Set $n$, $t$ as Equations~\eqref{eq:n}, \eqref{eq:t}, $W_1 = W$, and $\delta_1 = \delta$
\STATE Initialize the current learning task index $r_1 = 1$
\FOR{$c=1,2,..., n$}
\STATE Set $A_c$ to be an empty set
\STATE Set the ground truth labels of tasks $t_{r_c}$ to $t_{r_c+(\lfloor \frac{t}{|W_c|} \rfloor)}$ as $G_c = [g_{1,c}, g_{2,c}, ... , g_{\lfloor \frac{t}{|W_c|} \rfloor, c}]$
\STATE Set the historical accuracy of $W_c$ as $H_c = \{h_1, h_2,..., h_{|W_c|}\}$
\STATE Set the historical task number of $W_c$ as $N_c = \{n_1, n_2,..., n_{|W_c|}\}$
\FOR{each $w_i \in W_c$}
\STATE Assign learning tasks $t_{r_c}$ to $t_{r_c+(\lfloor \frac{t}{|W_c|} \rfloor)}$ to $w_i$ in batches, reveal the correct answers after $w_i$ submits
\STATE Get the answers $a_{i,c}$ of $w_i$ store to $A_c$ 
\ENDFOR
\STATE $r_{c+1} = r_{c}+(\lfloor \frac{t}{|W_c|} \rfloor) $
\STATE The predicted accuracy $p_c = \text{CPE}(A_c, G_c, H_c, |W_c|)$
\STATE The updated predicted accuracy with learning gains $\hat{p}_c = \text{LGE}(W_{c}, H_c, N_c, p_1, p_2, ..., p_{c})$
\STATE $W_{c+1} = \text{ME}(\hat{p}_c, W_c)$, $\delta_{c+1} = \frac{\delta_c}{2}$
\ENDFOR
\STATE Set $W_T$ to be the top $k$ workers with highest $\hat{p}_n$ in $W_{n+1}$. If $|W_{n+1}| < k$, set $W_T$ to be the top $k$ workers with highest $\hat{p}_{n-1}$ in $W_{n}$.
\RETURN $W_T$
\end{algorithmic}
\end{algorithm}

\section{Experiments} \label{sec:exp}
\subsection{Datasets} \label{sec:dataset}
Currently, no publicly available dataset records both the cross-domain worker historical profiles and the worker training process. Therefore, we have to construct new datasets that cover the two aspects of information to evaluate the performance of our method and the baselines. To this end, we build real-world and synthetic datasets, summarized in Table~\ref{tab:datasets}. We denote the number of learning tasks per batch as $Q$. The total budget $B = \lceil \log(\frac{|W|}{k}) \rceil*Q*|W|$, \rev{$\#\text{ of batches}=2^{\lceil\log\frac{|W|}{k}\rceil}-1$. Note that $Q$ and $k$ are the independent variables, while \# of batches and $B$ are dependent variables. We generate different synthetic datasets with different $|W|$ to study the influence of the size of the worker pool.}

\begin{table}[t!]
\setlength{\tabcolsep}{8pt}
  \caption{Dataset statistics}
  \begin{center}
  \vspace{-1em}
  \begin{tabular}{cccccc}
\hline
    \textbf{Datasets} & \textbf{$|$W$|$} & \textbf{Q} & \textbf{k} & \textbf{total \# of batches} & \textbf{B}\\
\hline
    RW-1 & 27 & 10 & 7 & 3 & 540\\
    \rev{RW-2} & \rev{35} & \rev{10} & \rev{9} & \rev{3} & \rev{700}\\
    S-1 & 40 & 20 & 5 & 7 & 2400\\
    S-2 & 50 & 20 & 5 & 7 & 3000\\
    S-3 & 80 & 20 & 5 & 15 & 6400\\
    S-4 & 160 & 20 & 5 & 31 & 16000\\
\hline
    \label{tab:datasets}
\end{tabular}
\end{center}
  \vspace{-3em}
\end{table}

\noindent \rev{\textbf{Real-world datasets: }We invited 27 and 35 workers to complete the Qualtrics survey~\cite{qualtrics-web} through volunteer recruitment and gMission~\cite{chen2014gmission} and denoted as RW-1 and RW-2 datasets. The tasks are Yes/No questions regarding image classification on three prior domains and one target domain. We chose Yes/No questions since many complex question types such as MCQs can be transformed from them~\cite{green1979multiple,dudley2006multiple}. We present the detailed information of the two real-world datasets in Table~\ref{tab:datasource}. Specifically, RW-1 examines workers' prior domain knowledge of animals (elephant and clownfish) and machines (plane) and evaluates their performance on plants (petunia). The key features that workers need to distinguish petunias from other flowers are color and shape. We also include RW-2 as a complement to RW-1: RW-2 focuses on finer-grained domains where the Peruvian lily, English marigold, and Lenten rose are all flowers. The key features that workers need to focus on differ: Peruvian lilies can be distinguished based on their color, while English marigolds and Lenten roses require detailed observation of petal and stamen shapes. By conducting experiments on both RW-1 and RW-2, we can comprehensively evaluate the performance and robustness of our approach and obtain interesting insights into cross-domain worker training.}
On each prior domain, each worker is asked to complete two batches of tasks. Each batch consists of 5 learning tasks and 5 working tasks. The answers are recorded to form the historical profiles of the workers. In the target domain, each worker needs to answer 30 learning tasks and 30 working tasks for us to record the worker training process. The learning and working tasks are assigned to each worker in batches. In each batch, workers are required to complete 10 learning tasks first, check the ground truth answers of the learning tasks, and then complete 10 working tasks. The learning tasks are used to train the workers gradually, while the working tasks are applied to test workers' annotation performance in the target domain. A sample learning task is shown in Figure~\ref{fig:problem}. Only the answers to the learning tasks are used as the algorithm input. The answers to the working tasks are used to evaluate the performance.

\begin{table}[t]
\setlength{\tabcolsep}{2.6pt}
  \caption{\rev{Details of real-world datasets}}
  \begin{center}
  \vspace{-1em}
  \begin{tabular}{lllll}
\hline
    \textbf{Dataset} & \textbf{Domain} & \textbf{Features} & \textbf{Knowledge} & \textbf{Sources}\\
\hline
    RW-1 prior-1 & Elephant & Color, Shape & Animal & \cite{elephant} \\
    RW-1 prior-2 & Clownfish & Color, Shape & Animal & \cite{clownfish,boom2012long,boom2012supporting} \\
    RW-1 prior-3 & Plane & Size & Machine & \cite{maji2013fine} \\
    RW-1 target & Petunia & Color, Shape & Plant & \cite{nilsback2008automated} \\
\hline
    RW-2 prior-1 & Peruvian lily & Color & Plant & \cite{nilsback2008automated} \\
    RW-2 prior-2 & Red fox & Shape & Animal & \cite{deng2009imagenet} \\
    RW-2 prior-3 & English marigold & Shape & Plant & \cite{nilsback2008automated} \\
    RW-2 target & Lenten rose & Shape & Plant & \cite{nilsback2008automated} \\
\hline
\end{tabular}
\end{center}
\vspace{-2em}
\label{tab:datasource}
\end{table}

\noindent \textbf{Synthetic datasets: }We further constructed synthetic datasets based on the distribution of the RW-1. We considered the synthetic datasets with worker pool sizes of $40, 50, 80$, and $160$ to simulate the different supply conditions. We set the number of learning tasks per batch on primal domains and the target domain to 10 and 20, respectively, on S-1, S-2, S-3, and S-4 datasets. We started by modeling the relationship among the four domains with a truncated multivariate normal distribution $N(\mu, \Sigma)$ within $(0,1)$, where the mean and standard deviation of the three prior domains are computed based on the learning task result of the workers on the three corresponding domains, while the mean and standard deviation of the target domain is calculated based on the first batch learning task results in the RW-1 dataset. The correlation parameters shown in Equation~\eqref{eq:init} are uniformly random initialized within $(0,1)$. Each synthetic worker was sampled from $N$ as $[h_1,h_2,h_3,h_T]^\mathsf{T}$, where $h_T \in (0,1)$ denotes the probability that the worker answers the target domain tasks correctly. We can thus obtain the annotation accuracy on the target domain learning tasks with the following \textit{answering rule: randomly select a number $x$ in $(0,1)$ if $x<h_T$, then the worker answers correctly. Otherwise, the worker answers wrongly.} We obtained the annotation accuracy of synthetic workers on the first batch of learning tasks and applied the modified IRT model in Equation~\eqref{eq:IRTmodi} to get the learning parameter $\alpha_i$ for each worker. Then we updated each worker's $h_T$ after each batch based on the modified IRT model with each worker's $\alpha_i$ and the annotation accuracy generated with the \textit{answering rule}. The top-k high-quality workers were selected based on the value of $h_T$ in the last batch.

\begin{table}[t]
\setlength{\tabcolsep}{5pt}
  \caption{\rev{Mean and standard deviation of RW-1 and synthetic datasets}}
  \begin{center}
  \begin{tabular}{lllll}
\hline
    \textbf{Dataset} & \textbf{Prior 1} & \textbf{Prior 2} & \textbf{Prior 3} & \textbf{Target}\\
\hline
    RW-1 & (0.70, 0.22) & (0.88, 0.10) & (0.58, 0.25) & (0.55, 0.17)\\
    S-1 & (0.72, 0.23) & (0.86, 0.13) & (0.53, 0.29) & (0.49, 0.18)\\
    S-2 & (0.64, 0.27) & (0.83, 0.15) & (0.51, 0.25) & (0.51, 0.20)\\
    S-3 & (0.66, 0.26) & (0.87, 0.13) & (0.54, 0.27) & (0.50, 0.18)\\
    S-4 & (0.68, 0.25) & (0.87, 0.13) & (0.54, 0.27) & (0.50, 0.18)\\
\hline
\end{tabular}
\end{center}
\vspace{-2em}
\label{tab:datadistribution}
\end{table}

\noindent \rev{\textbf{Consistency: }Notice that the synthetic datasets are generated based on RW-1, we now study the consistency between their distributions. As shown in Table~\ref{tab:datadistribution}, the computed mean and standard deviation of the multivariate normal distributions for RW-1 and synthetic datasets are close in the target and prior domains. Besides, we present the distribution of workers' annotation accuracy on the target domain for RW-1, S-1, S-2, S-3, and S-4 in our technical report~\cite{technical-report}. The real-world dataset RW-1 and the four synthetic datasets generated have similar distributions on the target domain. Specifically, we bucket the annotation accuracy, compute the Pearson correlations between RW-1 and each synthetic dataset, and find that all Pearson correlations $\rho$'s are larger than 0.75, which validates the consistency.} 

\begin{table*}[hbtp]
\setlength{\tabcolsep}{9pt}
  \caption{Experiment results}
  \vspace{-1em}
  \begin{center}
  \begin{tabular}{lcccccc}
\hline
    & \textbf{RW-1} & \rev{\textbf{RW-2}} & \textbf{S-1} & \textbf{S-2} &\textbf{S-3} &\textbf{S-4}\\
\hline
    \textbf{US~\cite{cao2015top,even2006action}} & 0.764 (4.5\% $\uparrow$) & \rev{0.956 (0.5\% $\uparrow$)} & 0.765 (8.5\% $\uparrow$) & 0.775 (6.8\% $\uparrow$) & 0.815 (4.3\% $\uparrow$) & 0.865 (2.4\% $\uparrow$)\\
    \textbf{ME~\cite{cao2015top,even2006action}} & 0.771 (3.5\% $\uparrow$) & \rev{0.944 (1.8\% $\uparrow$)} & 0.720 (15.3\% $\uparrow$) & 0.785 (5.5\% $\uparrow$) & 0.795 (6.9\% $\uparrow$) & 0.880 (0.7\% $\uparrow$) \\
    \textbf{Li et al.~\cite{li2014wisdom}} & 0.771 (3.5\% $\uparrow$) & \rev{0.936 (2.7\% $\uparrow$)} & 0.780 (6.4\% $\uparrow$) & 0.805 (2.9\% $\uparrow$) & 0.845 (0.6\% $\uparrow$) & 0.870 (1.8\% $\uparrow$) \\
\hline
    \textbf{ME-CPE} & 0.781 (2.2\% $\uparrow$) & \rev{0.950 (1.2\% $\uparrow$)} & 0.785 (5.7\% $\uparrow$) & 0.790 (4.8\% $\uparrow$) & 0.838 (1.4\% $\uparrow$) & 0.875 (1.3\% $\uparrow$) \\
\hline
    \textbf{Ours} & \textbf{0.798} & \rev{\textbf{0.961}} & \textbf{0.830} & \textbf{0.828} & \textbf{0.850} & \textbf{0.886} \\
\hline
    \textbf{Ground Truth} & 0.914 & \rev{1.000} & 0.885 & 0.875 & 0.915 & 0.975 \\
\hline
\end{tabular}
\end{center}
\label{tab:exp}
\vspace{-2em}
\end{table*}

\subsection{Baselines} 
In our experiment, we compared our proposed method with the general worker selection algorithms Median Elimination (ME), Uniform Sampling (US) discussed by~\cite{even2006action, cao2015top} and Li et al.'s method~\cite{li2014wisdom}. We chose these baselines because they do not require additional social interaction information (required by~\cite{zhao2017context,wu2021task}) and are comparable in terms of tasks and goals (Liu et al.~\cite{liu2013scoring} focus on optimizing the number of golden questions used, while the worker selection algorithm used is US~\cite{even2006action}; Yadav et al.~\cite{yadav2022multi} aim at forming high-performance worker teams, which is not comparable under our problem setting). We introduce the baselines as follows:  
\begin{itemize}[leftmargin=*]
    \item Uniform Sampling (US)~\cite{even2006action, cao2015top}: Assign each worker the same amount of learning tasks and select the top-k workers that have the highest accuracy.
    \item Median Elimination (ME)~\cite{even2006action, cao2015top}: Assign each worker $\frac{t}{|W_c|}$ learning tasks in round $c$, after each round, apply Algorithm~\ref{algo:ME} to select the best half of workers.
    \item Li et al.~\cite{li2014wisdom}: Adopt linear regression on the multiple features of workers and then select workers based on the regressed values. In our experiment, we use the historical profiles as the features for the regression process.
\end{itemize}

US and ME focus on selecting high-quality workers based on their annotation performance in the worker training process, while the Li et al. approach focuses on employing the historical profile features to identify high-quality workers.

\subsection{Experiment Setting} \label{sec:exp_setting}

To ensure fairness, we allocated the same amount of budget for our method and the baselines. The methods' performances are evaluated with respect to the average annotation accuracy of the selected workers on the target domain working tasks in the last round. For example, in \rev{RW-1 and RW-2 datasets}, the worker medium elimination process terminates in two rounds. We used the annotation accuracy on the working tasks in the second round as the performance criteria. The difficulty parameters of prior domains $\beta_d$ were initialized as $\beta_d = \ln(\frac{1}{a_d}-1)$, where $a_d$ is the averaged annotation accuracy for all the real workers on the domain $d$. For the target domain, we set $\beta_T=0$ so that Equation~\eqref{eq:IRTmodi} would be $0.5$ when $K_j=0$. Since the tasks are Yes/No questions, we believe that $a_T=0.5$ is a good choice when no prior knowledge regarding the target domain tasks is given. We further conducted a parameter sensitivity experiment in Section~\ref{sec:para} to validate our choice. We adopted the same initialization setting regarding the difficulty parameters for the synthetic datasets. The initial multivariate normal distribution $N(\mu, \Sigma)$ was generated as follows: $\mu_1, \mu_2, ..., \mu_D$ and $\sigma_1, \sigma_2,..., \sigma_D$ were initialized based on the mean and variance of workers' annotation accuracy on the corresponding prior domains; $\mu_T$ was initialized to $0.5$, and $\sigma_T$ was initialized as $\frac{1}{D}\sum_{d=1}^D \sigma_d$. The correlation parameters were uniformly random initialized in the $(0,1)$ range. The gradients with respect to $\Bar{\mu}$ and $\Bar{\Sigma}$ in the log-likelihood function $L$ (Equation~\eqref{eq:MLE}) was computed based on backpropagation~\cite{werbos1990backpropagation}. We set the learning rates used for gradient descent as $r_1=1e-7$, $r_2=1e-4$, and the epochs for gradient descent as $G=50$. The average accuracy of each method was recorded. The implementation and data are available at \url{https://github.com/ysunbp/crowdsourcing}.

\subsection{Experiment Results}

We present the experiment results regarding the best-k worker average annotation accuracy and the relative improvements of our method over baselines in Table~\ref{tab:exp}. The ground truth results are presented in the bottom line of Table~\ref{tab:exp}. In general, we observe that our method performs the best baseline approach on real-world and synthetic datasets. Specifically, on \rev{RW-1 and RW-2 datasets}, our method outperforms \rev{US~\cite{cao2015top,even2006action} by $4.5\%$ and $0.5\%$, while boosting the performance of ME~\cite{cao2015top,even2006action} by $3.5\%$ and $1.8\%$, and Li et al.~\cite{li2014wisdom} by $3.5\%$ and $2.7\%$, respectively.} On the four synthetic datasets, our method outperforms US, ME, and Li et al. by $5.5\%$, $7.1\%$, and $2.9\%$ on average. We attribute the performance uplift of our method over US and ME to the fact that we additionally consider the cross-domain historical profile information of workers and apply proper simulation of the learning gain of workers during the worker training process. The reason why our approach outperforms Li et al. can be attributed to the appropriate worker elimination process applied and the proper simulation of the learning gain of workers. We further notice that the average relative performance uplifts of our approach over the three baselines are $10.1\%$, $5.1\%$, $3.9\%$, and $1.6\%$, which decrease as the number of workers increases. We attribute this phenomenon to the fact that the number of high-performance workers increases as the size of the worker pool gets large. As a result, the accuracy difference induced by different worker selection strategies of different approaches is likely to decrease, and thus the performance uplift becomes smaller as the size of the worker pool increases. Overall, our method performs persistently well on both real-world and synthetic datasets, which implies the effectiveness and robustness of our approach for cross-domain worker selection.

\begin{figure}[t]
  \centering
  \includegraphics[width=0.7\linewidth]{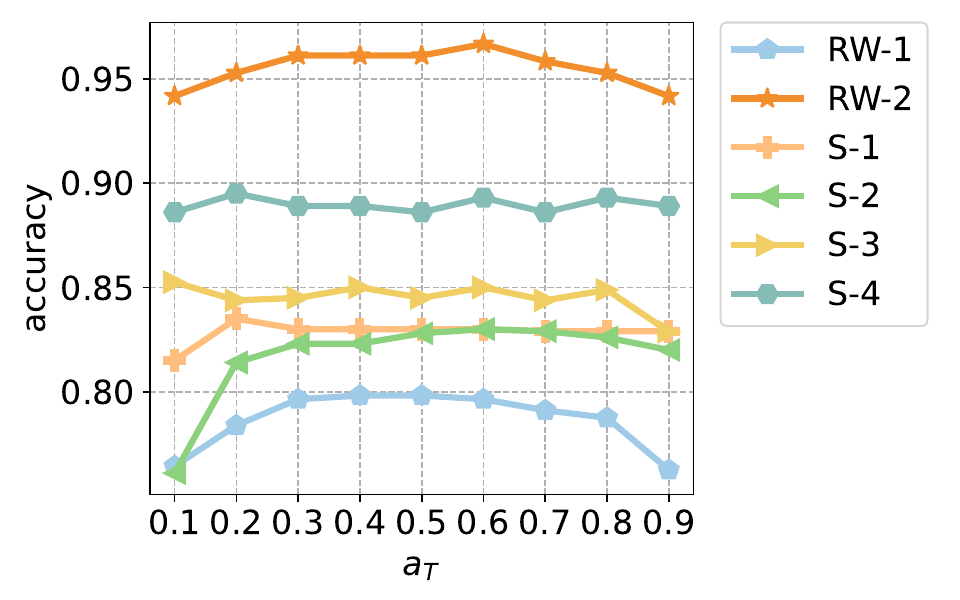}
  \vspace{-1em}
  \caption{\rev{The sensitivity analysis regarding the initialized annotation accuracy of the target domain: $a_T = \frac{1}{1+e^{\beta_T}}$ on different datasets. }}
  \label{fig:parasense}
    \vspace{-2em}
\end{figure}

\begin{figure*}[t]
\centering 
  \subfloat[Value of k (RW-1 dataset)]{
\label{fig:RW1-k}
  \includegraphics[width=0.25\linewidth]{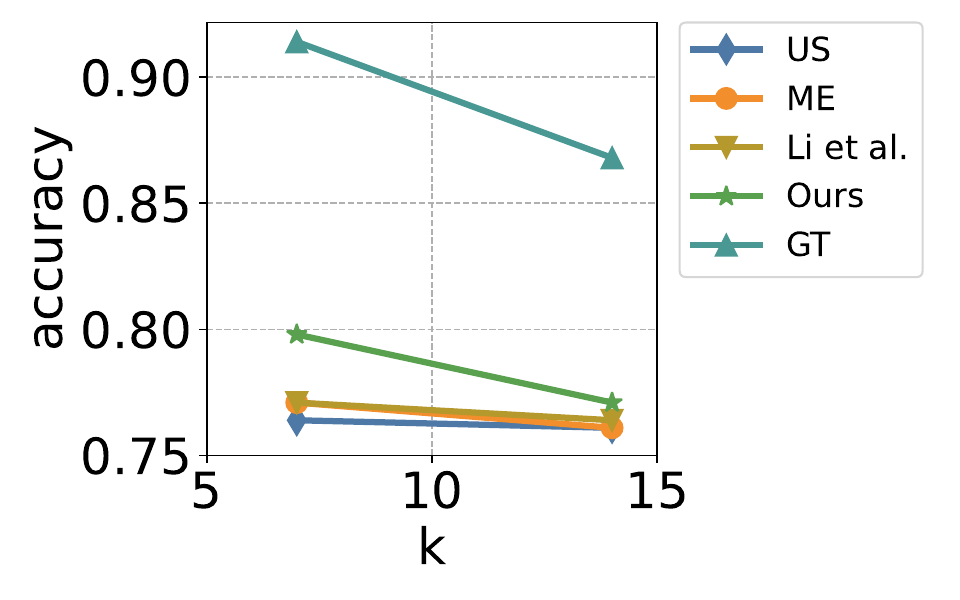}}
  \subfloat[\rev{Value of k (RW-2 dataset)}]{
\label{fig:RW2-k}
  \includegraphics[width=0.25\linewidth]{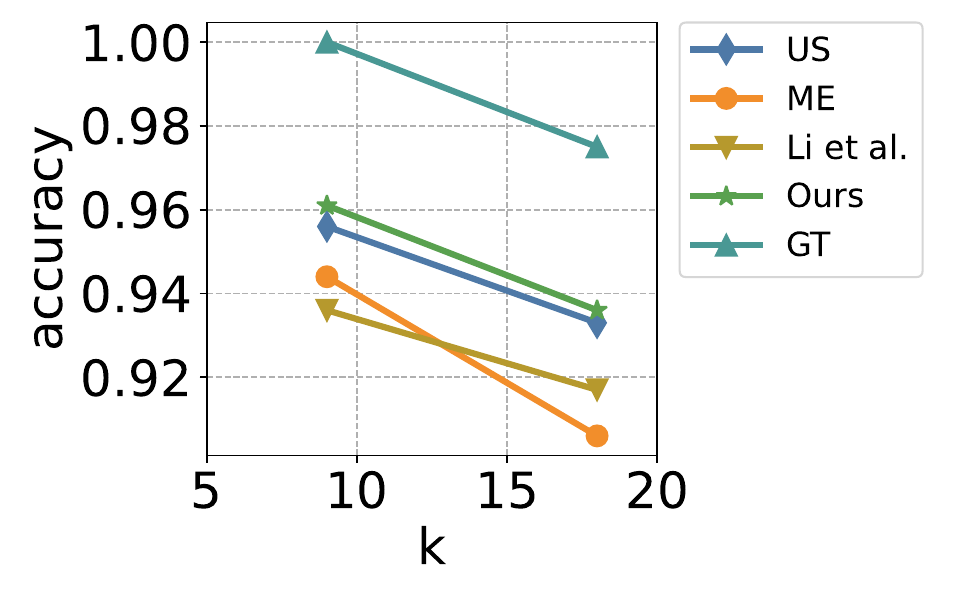}}
  \subfloat[Value of k (S-1 dataset)]{
  \label{fig:S1-k}
  \includegraphics[width=0.25\linewidth]{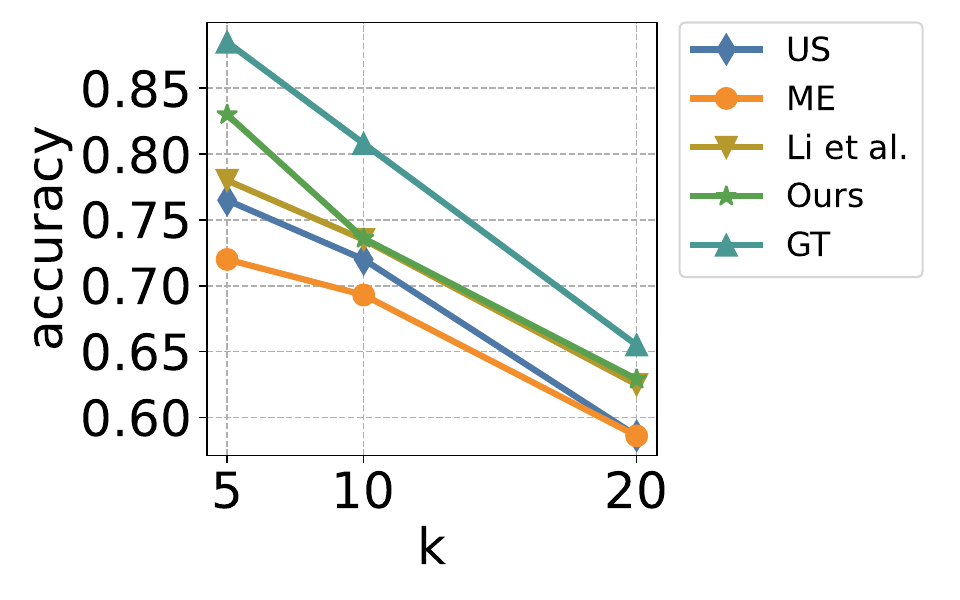}}
  \\
  \subfloat[Value of k (S-2 dataset)]{
  \label{fig:S2-k}
  \includegraphics[width=0.25\linewidth]{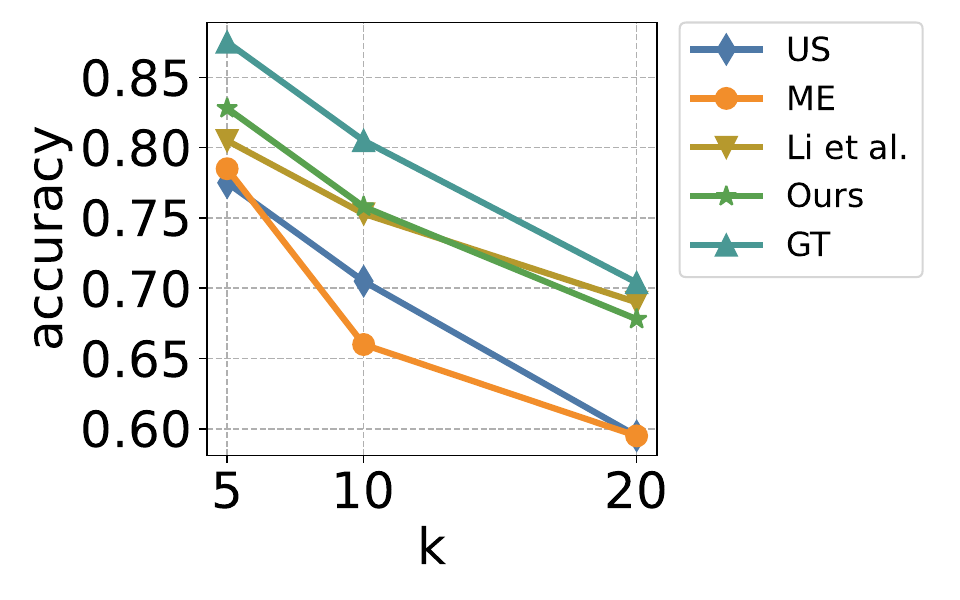}}
  \subfloat[Value of k (S-3 dataset)]{
  \label{fig:S3-k}
  \includegraphics[width=0.25\linewidth]{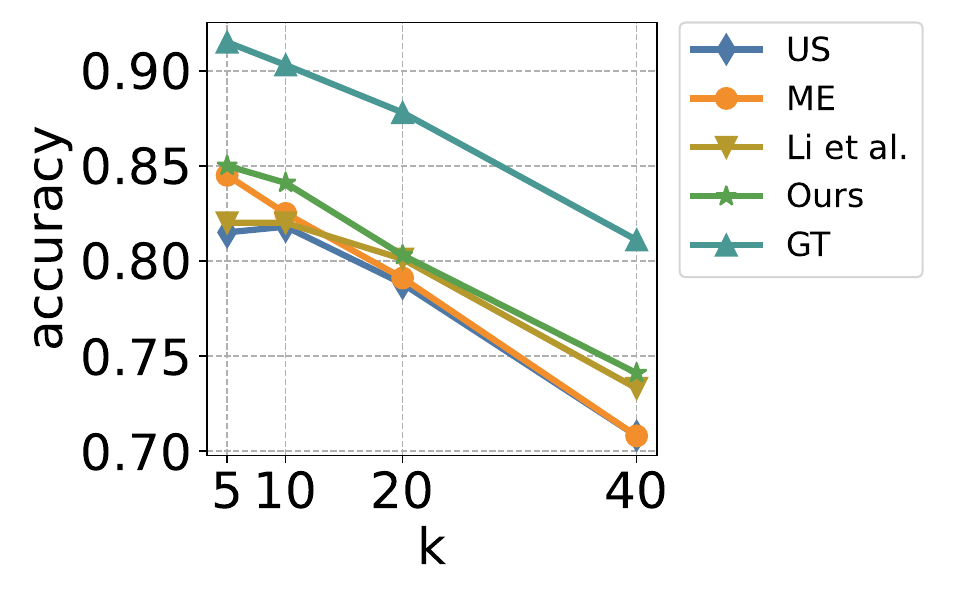}}
  \subfloat[Value of k (S-4 dataset)]{
  \label{fig:S4-k}
  \includegraphics[width=0.25\linewidth]{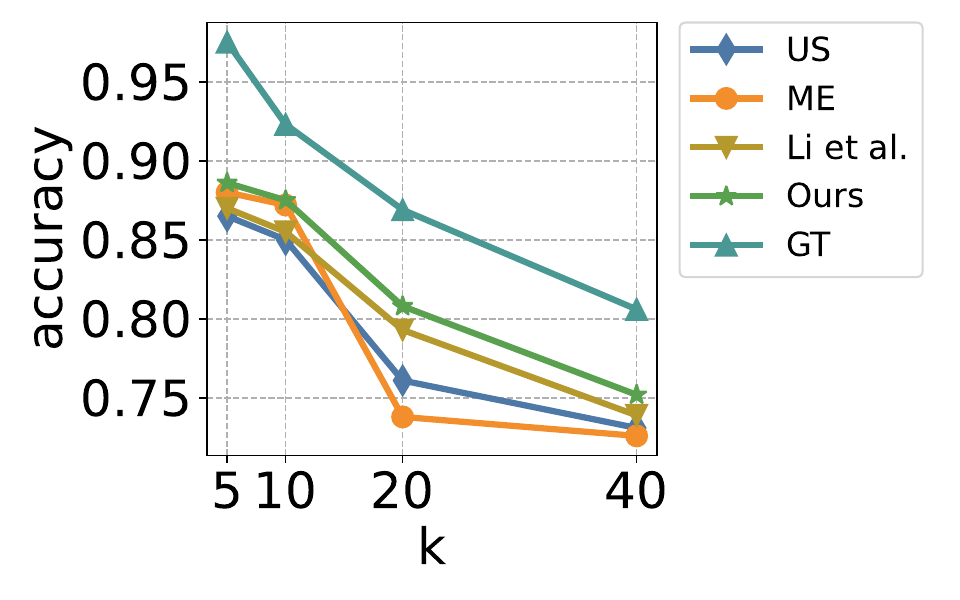}}
  \caption{Parameter sensitivity experiments of the number of selected workers.}
  \label{fig:para-k}
  \vspace{-2em}
\end{figure*}

\subsection{Ablation Study}
We conducted an ablation study regarding the following variants to understand the mechanism of different components:
\begin{itemize}[leftmargin=*]

    \item ME~\cite{cao2015top, even2006action}: ME is the backbone of our method, where the CPE and LGE (Worker Quality Estimation) are removed.
    \item ME-CPE: ME-CPE is a variant of our method where the LGE component is removed.
\end{itemize}

We present the experiment results in Table~\ref{tab:exp}. We compare ME~\cite{cao2015top, even2006action} with ME-CPE to demonstrate the effect of the Cross-domain-aware Performance Estimation. The comparison between ME-CPE and our method shows the influence of the Learning Gain Estimation. First, we note that ME-CPE outperforms ME by $1.3\%$ and \rev{$0.6\%$} and our method further boosts the performance of ME-CPE by $2.2\%$ and \rev{$1.2\%$} on \rev{RW-1 and RW-2 datasets}. This shows the effectiveness brought by CPE and LGE in estimating worker quality. CPE helps the algorithm capture the cross-domain information to estimate the annotation ability of workers, while LGE captures the learning gain of workers during the worker training process. On S-1, S-2, S-3, and S-4 datasets, our method improves over ME-CPE by $5.7\%$, $4.8\%$, $1.4\%$, and $1.3\%$ respectively. We attribute the performance improvement of our method over ME-CPE to the LGE component, which obtains a more accurate estimation of workers' performance in the target domain during training. ME-CPE outperforms ME by relatively $9.0\%$, $0.6\%$, and $5.4\%$ on S-1, S-2, and S-3 datasets while performing slightly worse than ME on the S-4 dataset. On average, ME-CPE relatively improves the performance of ME by $3.6\%$ on the four synthetic datasets. In general, ME-CPE can improve or achieve comparable performance as ME. The CPE component can effectively capture the cross-domain information, which is helpful for identifying high-quality workers.

\subsection{Method Parameter Sensitivity} \label{sec:para}

We analyzed the impact of the critical parameter of our method $a_T$, which is the initialized annotation accuracy of the target domain and is related to the initialization of the difficulty parameter $\beta_T$ ($a_T = \frac{1}{1+e^{\beta_T}}$). Figure~\ref{fig:parasense} presents the results. We notice that the performance of our method is relatively stable when the value of $a_T$ is set within the range $[0.2,0.8]$. In practice, we suggest setting $a_t$ according to the difficulty level of the tasks. If the tasks are relatively easy, a large value of $a_T$ can be adopted. Otherwise, a small value of $a_T$ should be considered. If no domain knowledge regarding the difficulty level is available, we can set the value of $a_T$ based on the nature of the tasks. Since our tasks are Yes/No quuestions, a natural choice of $a_T$ would be $0.5$. The sensitivity analysis coincides with our selection of $a_T$ in Section~\ref{sec:exp_setting}: our method achieves a stable and good performance when $a_T=0.5$.

\subsection{Dataset Parameter Sensitivity}
To comprehensively compare the performance of our method and baselines, we further conducted experiments regarding the important dataset parameters: the number of selected workers $k$ and the number of learning tasks per batch $Q$. Since the number of learning tasks per batch cannot be changed once the RW datasets are collected, we conducted experiments on the four synthetic datasets to analyze its effect.

The value of $k$ determines the number of rounds $n$ required to obtain the top-k workers. As shown in Table~\ref{tab:datasets}, the values of $k$ used in our main experiments are \rev{7 and 9 for the RW-1 and RW-2 datasets} and 5 for the four synthetic datasets, which leads to 2 rounds for \rev{RW-1 and RW-2 datasets}, 3 rounds for the S-1 and S-2 datasets, 4 rounds for the S-3 dataset, and 5 rounds for the S-4 dataset. In the parameter sensitivity experiment presented in this section, we further increased the value of $k$ to 14 and \rev{18 for RW-1 and RW-2 datasets}, 10 and 20 for the S-1 and S-2 datasets, and 10, 20, and 40 for the S-3 and S-4 datasets. The reason for experimenting with the increased number of $k$ is to present a comprehensive view of the performance of our approach from the beginning stage of worker selection (when the value of $k$ is large) to the ending stage of worker selection (when the value of $k$ is small). For example, when changing the value of $k$ from 14 to 7 on the RW-1 dataset, we can analyze the performance change of our method when conducting one round and two rounds of eliminations. \rev{As shown in Figures~\ref{fig:RW1-k} and~\ref{fig:RW2-k}, on RW-1 and RW-2 datasets, our method consistently outperforms all the baseline approaches, when we increase the value of $k$.} On the synthetic datasets S-1, S-3, and S-4, as shown in Figures~\ref{fig:S1-k}, \ref{fig:S3-k}, and \ref{fig:S4-k}, our approach still outperforms all the baselines when the value of $k$ increases. On the S-2 dataset (Figure~\ref{fig:S2-k}), our approach outperforms all the baselines when the value of $k$ is set to 5 and 10, while is slightly worse than the performance of Li et al.~\cite{li2014wisdom} when the value of $k$ further increase to 20. We further observe that on the RW-1 dataset, when $k=14$, our approach and Li et al. have similar performance. Similar phenomena can also be observed on the S-1, S-3, and S-4 datasets when we set the value of $k$ to 20, 40, and 40. We attribute this to the fact that when the value of $k$ is large (i.e., the model is at the beginning stage of elimination), the long-term learning improvement of the workers on the target domain is not yet significant, the linear regression approach introduced by Li et al.~\cite{li2014wisdom} can capture the static cross-domain information. However, as the value of $k$ decreases (i.e., the elimination process proceeds), the dynamic cross-domain performance estimation and the learning gain estimation help our approach to outperform Li et al.~\cite{li2014wisdom}.

As for the number of learning tasks $Q$, the default value is 20. We changed the value of $Q$ to 16, 30, and 40, while keeping the value of $k$ unchanged with a changing total budget $B$ in this section to evaluate its influence and presented the experimental results in Figure~\ref{fig:para-Q}. We first notice that our approach consistently outperforms all baselines on four synthetic datasets with different $Q$. We further observe that on four synthetic datasets, the performance of our approach and baselines tends to get close when $Q$ increases. We attribute this phenomenon to the fact that when the budget is arbitrarily large, the improvement brought by adopting cross-domain information is reduced since the algorithm can get an accurate estimation of workers' target domain knowledge based on a large amount of learning tasks assigned to workers. However, when the total budget is small, our approach efficiently utilizes the cross-domain information to boost the worker selection performance of ME. In real-world applications, selecting workers effectively with a relatively small number of learning tasks is crucial, since in reality, the ground truth answers of golden questions in each domain require manual collection and thus are hard to obtain. In this sense, the performance uplift of our approach over other baselines when $Q$ is small is favored and useful for real-world applications. 

\begin{figure*}[t]
\centering
  \subfloat[Learning tasks Q (S-1 dataset)]{
\label{fig:learning-S1}
  \includegraphics[width=0.24\linewidth]{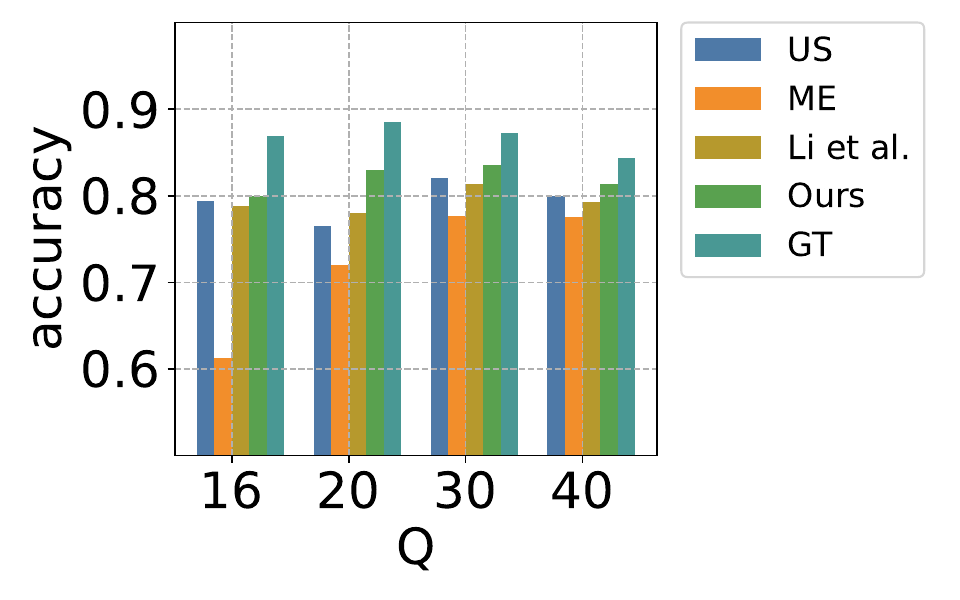}}
  \subfloat[Learning tasks Q (S-2 dataset)]{
\label{fig:learning-S2}
  \includegraphics[width=0.24\linewidth]{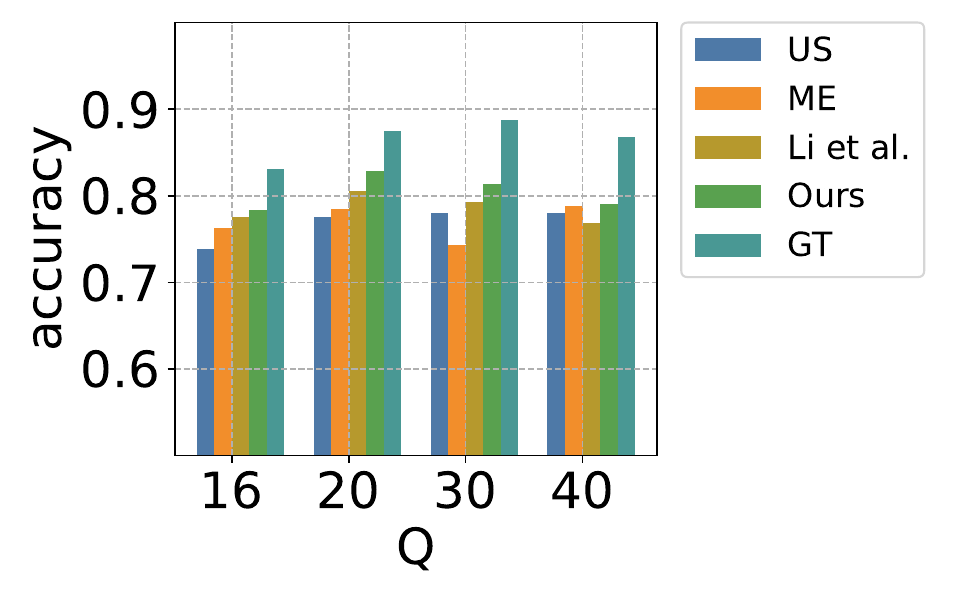}}
  \subfloat[Learning tasks Q (S-3 dataset)]{
\label{fig:learning-S3}
  \includegraphics[width=0.24\linewidth]{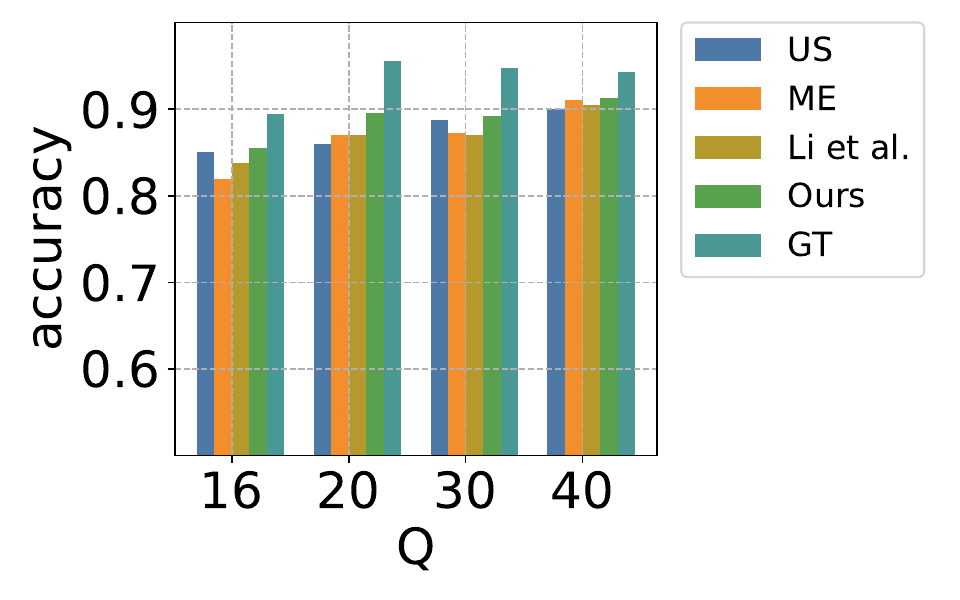}}
  \subfloat[Learning tasks Q (S-4 dataset)]{
\label{fig:learning-S4}
  \includegraphics[width=0.24\linewidth]{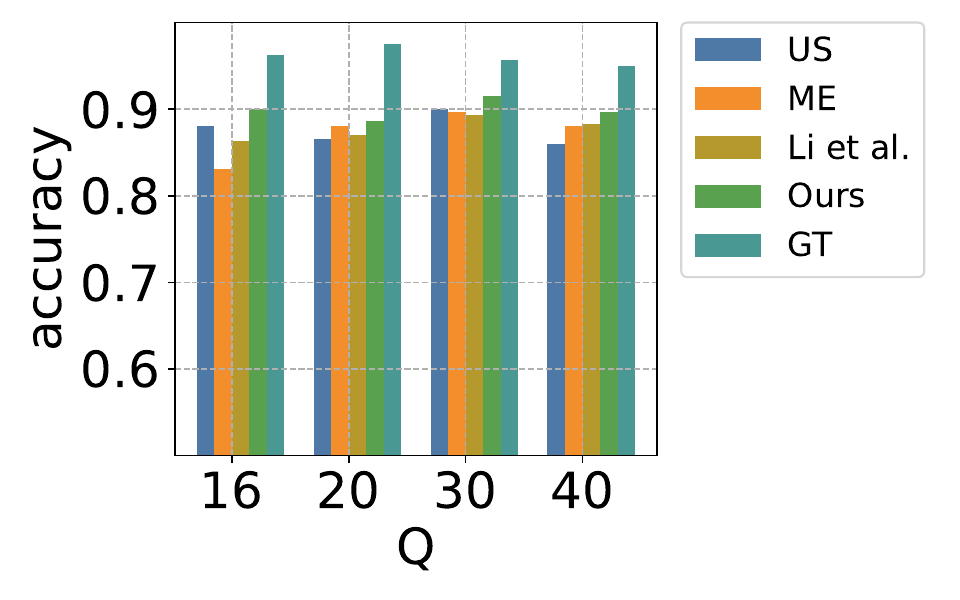}}
  \caption{The parameter sensitivity experiments of the number of learning tasks per batch $Q$.}
  \label{fig:para-Q}
\vspace{-2em}
\end{figure*}

\subsection{\rev{Discussion}}
\label{sec:runtime}
We recorded the running time results on one Intel Xeon Gold 6240 CPU @ 2.60GHz. \rev{Our method takes 3.9s, 5.0s, 6.3s, 7.8s, 13.4s, and 28.9s to select the best workers on the RW-1, RW-2, S-1, S-2, S-3, and S-4 datasets.} Compared with the \rev{median} completion times of our two surveys (\rev{1185s and 986s}), the running time cost of our method is acceptable. As suggested by~\cite{zheng2018dlta}, the average completion time of tasks on AMT usually takes hours; we believe that the running time can facilitate the needs for real-world crowdsourcing applications. 

\rev{The time used for workers to complete learning tasks inside the survey is approximately 500s for the two surveys. We observe that the average accuracy of all workers increases from 0.55 to 0.79 and from 0.65 to 0.85 on RW-1 and RW-2 respectively after a single round of worker training with 10 questions. Despite the additional training time required, we believe that by introducing worker training into the worker selection process of crowdsourcing, we can significantly improve the overall worker performance in the target domain. As for the cost of the worker training process: denote the number of learning tasks and working tasks assigned as $|T_l|$ and $|T_w|$, the annotation accuracy before and after the worker training as $a_t$ and $a_t'$. For simplicity, we consider the effect on one single worker with one round of training, assume the worker's accuracy is the same as the average accuracy of all workers, and consider the same monetary cost for completing each learning and working task. If $|T_w|/|T_l|>\frac{a_t}{a_t'-a_t}$, under the same total worker learning and working budget, the number of correctly annotated samples in $T_w$ for the worker with worker training process would be greater than that without worker training. In our case, once $|T_w|/|T_l|$ is greater than 2.3 and 3.3 for RW-1 and RW-2 respectively, then the additional monetary cost can be counteracted. Furthermore, our worker training phase interacts with the workers by revealing the correct answers to learning tasks to workers promptly. Throughout multiple worker training rounds, workers can learn about their overall improved performance in the target domain, have a sense of accomplishment, and obtain new skills related to the target domain. As discussed by previous works~\cite{dontcheva2014combining, de2021increasing, hackman1976motivation, tang1991effects}, with timely feedback received and new skills learned, workers tend to have improved engagement and performance. Therefore, we believe the worker training phase improves the engagement of workers, and stimulates workers to explore more useful features in the target domain.}

\rev{We further report the estimated correlation between domains on the RW-1 and RW-2 datasets. Specifically, the correlation parameters estimated by our method are $0.50$, $0.69$, and $0.65$ for Plane-Flower (P-F), Fish-Flower (F-F), and Elephant-Flower (E-F) on RW-1 and $0.23$, $0.10$, and $0.68$ for Peruvian lily-Lenten rose (P-L), Red fox-Lenten rose (R-L), and English marigold-Lenten rose (E-L) on RW-2. The correlation parameters for F-F and E-F are larger than that for P-F, which coincides with our intuition that workers sensitive to color and shape differences (good at fish and elephant domains) are likely to perform well in distinguishing flowers. The correlation parameters for P-L and E-L are larger than that for R-L, which means the workers who are good at distinguishing other flowers are good at distinguishing Lenten roses. Besides, the correlation for E-L is larger than P-L. To distinguish English marigold from its counter-parts, workers need to notice the small differences in petals and stamen (shape), which is close to the requirement of distinguishing Lenten rose; while to identify the Peruvian lily from its counter-parts, workers only need to pay attention to color difference. As a result, the correlation for E-L is larger than P-L. More details of the prior and target domains are presented in Table~\ref{tab:datasource}.}

\section{Conclusion} \label{sec:conclude}

In this paper, we formulated the cross-domain-aware worker selection with training problem and proposed a novel algorithm based on Medium Elimination to resolve it. Specifically, two estimation components CPE and LGE are designed to incorporate cross-domain knowledge information and capture the learning gains during worker training. Real-world and synthetic cross-domain-aware worker selection with training datasets were collected to evaluate different approaches. We conducted extensive experiments on real-world and synthetic datasets to show that our method outperforms all the state-of-the-art baselines on real-world and synthetic datasets. We confirm that applying CPE and LGE can capture cross-domain knowledge information and estimate the learning gain during the worker training process. As a future direction, we aim to construct a unified multi-domain taxonomy that optimizes the worker training and selection process.

\section*{Acknowledgment}
Jiachuan Wang and Libin Zheng are the corresponding authors. Peng Cheng's work is supported by the National Natural Science Foundation of China under Grant No. 62102149. Libin Zheng is supported by the National Natural Science Foundation of China No. 62102463 and the Natural Science Foundation of Guangdong Province of China No. 2022A1515011135. Lei Chen’s work is partially supported by National Key Research and Development Program of China Grant No. 2023YFF0725100, National Science Foundation of China (NSFC) under Grant No. U22B2060, the Hong Kong RGC GRF Project 16213620, RIF Project R6020-19, AOE Project AoE/E-603/18, Theme-based project TRS T41-603/20R, CRF Project C2004-21G, Guangdong Province Science and Technology Plan Project 2023A0505030011, Hong Kong ITC ITF grants MHX/078/21 and PRP/004/22FX, Zhujiang scholar program 2021JC02X170, Microsoft Research Asia Collaborative Research Grant and HKUST-Webank joint research lab grants.

\clearpage
\balance
\bibliographystyle{IEEEtranS}
\bibliography{IEEEabrv,conference_101719.bib}

\vspace{12pt}

\end{document}